\documentclass[manuscript,screen]{acmart}
\pdfoutput=1

\AtBeginDocument{%
  \providecommand\BibTeX{{%
    \normalfont B\kern-0.5em{\scshape i\kern-0.25em b}\kern-0.8em\TeX}}}

\usepackage{graphicx}
\graphicspath{{Figures/}}
\usepackage{subcaption}
\usepackage[ruled,vlined]{algorithm2e}
\usepackage[normalem]{ulem}
\usepackage{booktabs}
\usepackage{amsmath}

\definecolor{Gray}{gray}{0.92}
\definecolor{LightCyan}{rgb}{0.88,1,1}

\renewcommand\footnotetextcopyrightpermission[1]{} 
\acmJournal{TOIT}

\setcitestyle{numbers,sort&compress}

\begin{document}

\title{Anonymizing Sensor Data on the Edge: A Representation Learning and Transformation Approach}

\author{Omid Hajihassani}
\email{hajihass@ualberta.ca}
\author{Omid Ardakanian}
\email{ardakanian@ualberta.ca}
\affiliation{%
  \institution{University of Alberta}
  \city{Edmonton}
  \state{Canada}
}
\author{Hamzeh Khazaei}
\email{hkh@eecs.yorku.ca}
\affiliation{%
  \institution{York University}
  \city{Toronto}
  \state{Canada}
}
\renewcommand{\shortauthors}{O. Hajihassani et al.}

\begin{abstract}
The abundance of data collected by sensors in Internet of Things (IoT) devices, and the success of deep neural networks in uncovering hidden patterns in time series data have led to mounting privacy concerns. This is because private and sensitive information can be potentially learned from sensor data by applications that have access to this data. In this paper, we aim to examine the tradeoff between utility and privacy loss by learning low-dimensional representations that are useful for data obfuscation. We propose deterministic and probabilistic transformations in the latent space of a variational autoencoder to synthesize time series data such that intrusive inferences are prevented while desired inferences can still be made with sufficient accuracy. \textcolor{red}{In the deterministic case, we use a linear transformation to move the representation of input data in the latent space such that the reconstructed data is likely to have the same public attribute but a different private attribute than the original input data. In the probabilistic case, we apply the linear transformation to the latent representation of input data with some probability.} We compare our technique with autoencoder-based anonymization techniques and additionally show that it can anonymize data in real time on resource-constrained edge devices.
\end{abstract}



\maketitle

\fancyfoot{}
\thispagestyle{empty}

\section{Introduction}
Internet of Things (IoT) systems are becoming increasingly ubiquitous in our daily lives.
They utilize a variety of sensors, actuators, and processing units 
to monitor and control our surrounding environment and generate valuable insights.
The sensors embedded in these systems collect a large amount of time series data, 
including audio and video~\cite{noto2016contracting, noda2018google, jeon2018iot}, temperature~\cite{hernandez2014smart}, 
and inertial data~\cite{hassan2018robust}.
\textcolor{red}{This data usually contains important information about 
various attributes pertaining to the user and their local environment.
These attributes can be classified into public and private attributes,
where the distinction lies in whether inferring them from the sensor data is desired or approved by the user.
For example, human activity is deemed a public attribute by people who wear a fitness tracker,
but mental health can be considered a private attribute if they do not 
want it to be inferred from this data.
Similarly, electricity consumption is a public attribute of people who installed a smart meter at home,
while their activities of daily living are typically deemed a private attribute.}

Despite the many applications and services that rely on IoT devices, 
their proliferation can jeopardize user privacy.
This is because raw sensor data is often shared with third-party applications 
that may use it to make sensitive, unsolicited inferences about their private attributes.
Recent advances in deep learning and edge computing have made it easier 
to perform these inferences even at the edge, increasing the privacy risks.
\textcolor{red}{This calls for the design of low-cost techniques to obfuscate sensor data 
on IoT/edge devices before applications can access this data.
These techniques should ensure that existing applications can work 
on the obfuscated data without any modifications to their code.}


Extensive research has been done in recent years to enable privacy-preserving data analysis 
through $k$-anonymity~\cite{bayardo2005data, jia2017pad}, 
differential-privacy~\cite{dwork2008differential, phan2016differential}, or 
by applying various machine learning techniques~\cite{hajihass2020latent, malekzadeh19}. 
\textcolor{red}{However, most approaches that prevent sensitive inferences or reduce their accuracy
(a measure of \emph{privacy loss}),
greatly reduce the accuracy of desired inferences (a measure of \emph{utility}).
We believe that in practice a significant loss of utility is not an acceptable tradeoff for privacy.}
A promising approach should utilize the available data to its fullest potential without compromising user privacy.

Deep generative models, in particular autoencoders, have been used to obfuscate sensor data such that
the risk of disclosing private attributes is reduced, while ensuring that public attributes remain 
intact~\cite{malekzadeh19,liu2019privacy,Olympus2019}.
For example, in~\cite{malekzadeh19} autoencoders trained in an adversarial fashion 
are used to address the privacy issues. 
We call such approaches \textcolor{red}{\textit{adversarial model-based}} since 
they use \textcolor{red}{specific} inference models for adversarial training.
Unfortunately as shown in~\cite{hajihass2020latent}, 
they can only fool these models and are susceptible to an \emph{attribute inference attack}
in which a model is trained by the attacker to re-identify the private attribute that has been concealed.
\textcolor{red}{In this attack, the attacker passes a sufficient amount of sensor data with known labels, 
i.e., private attributes, through the anonymization network, 
and trains a model on the output of this network, i.e., the obfuscated data, along with the corresponding labels.
Once trained, this model can be used to re-identify the private attribute of any user,
thereby reversing the anonymization process.}





To address this problem, we propose a novel data anonymization approach 
that entails learning the latent representation of time series data 
through a Variational Autoencoder (VAE) and transforming this representation
in a certain way to conceal the private attribute.
\textcolor{red}{This approach does not require adversarial training using a specific model, 
hence we refer to it as \textit{adversarial model-free} anonymization.}
Specifically, it extends the variational autoencoder-based technique 
proposed in our previous work~\cite{hajihass2020latent} in two main ways. 
First, we change the loss function of the conventional VAE by introducing 
a term that accounts for the private attribute classification error.
This helps to learn a latent representation that correlates with the private attribute.
Second, we train a separate VAE for each public attribute class 
and select the appropriate VAE at runtime depending on the inferred public attribute.
This is necessary for training useful and compact autoencoders as we discuss later.
\textcolor{red}{Using this anonymization technique, private attributes can be obscured 
before sensor data is used by on-device or cloud applications for various analyses.}

\begin{figure}[t]
\centering
  \includegraphics[scale=.6]{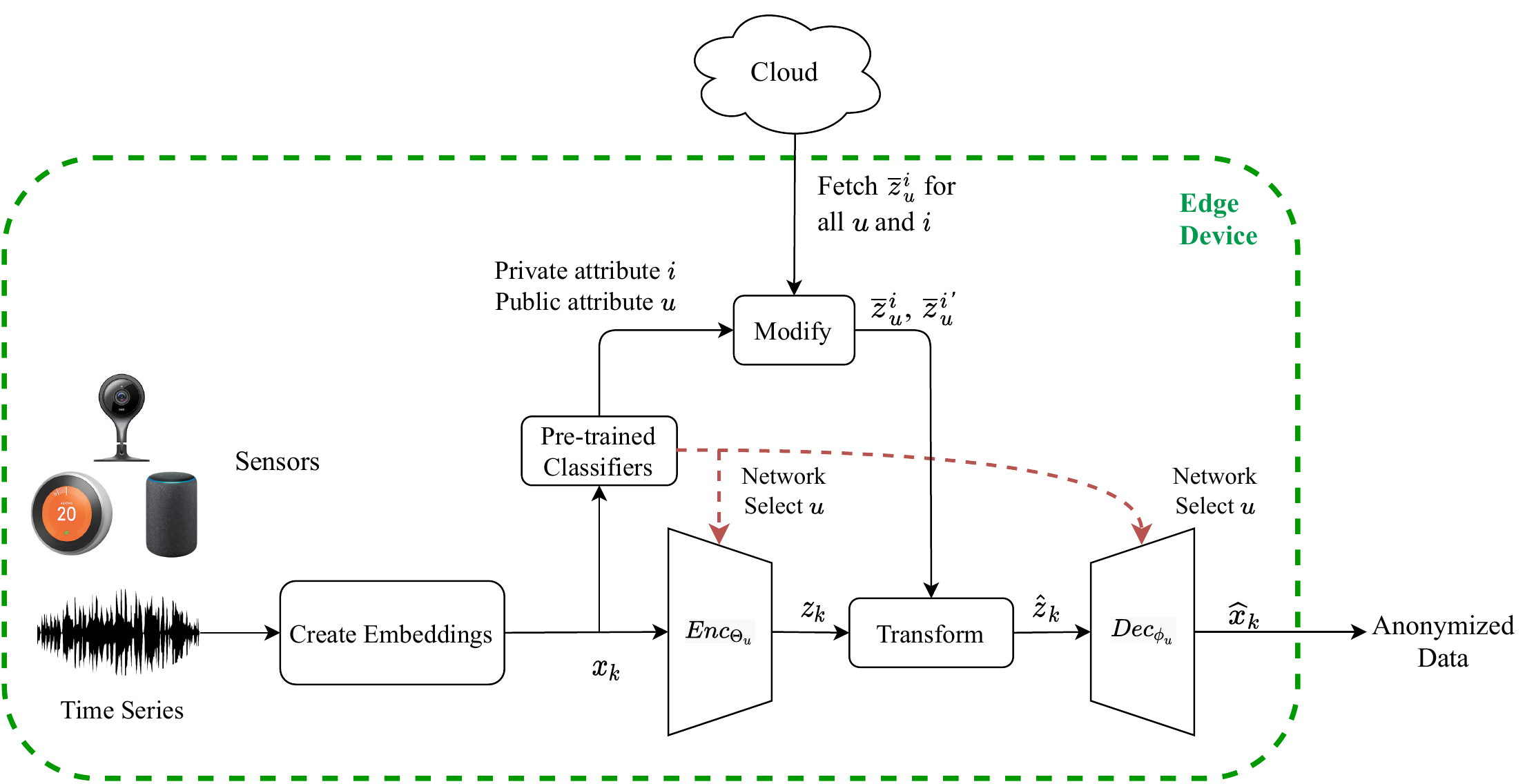}
  \caption{Data anonymization on an edge/IoT device using the proposed adversarial model-free anonymization technique. 
  The mean latent representation for each pair of private and public class labels is assumed 
  to be stored in a central (cloud) server. \textcolor{red}{Here $x_k$ is the $k^{\text{th}}$ data embedding, $z_k$ is its corresponding latent representation, $\hat{z}_k$ is the modified version of this latent representation, and $\hat{x}_k$ is the obfuscated version of the input data. 
  Moreover, $\bar{z}^i_u$ is the average value of all latent representations 
  that correspond to data with public attribute $u$ and private attribute $i$.}}
  \label{fig:modvae}
\end{figure}

\textcolor{red}{Figure~\ref{fig:modvae} shows different components of the proposed anonymization technique, 
including the attribute-specific VAE and transformation performed in the latent space of a VAE.
This anonymization technique runs on the IoT/edge device.
We assume a central server, which can be in the cloud, 
periodically broadcasts the average latent representations 
for all possible combinations of public and private attributes 
to the edge devices that have adopted this anonymization technique.
In each edge device, we create an embedding of time series data generated by one or multiple sensors
and predict the private and public attributes of each embedding using pretrained classifiers.
The predicted attributes are used to select the appropriate VAE and to 
load the average latent representations that correspond to the identified public attribute.
The average latent representations are then used to modify the latent representation 
of the input data that is produced by the encoder in a deterministic or probabilistic fashion.
These two \emph{mean manipulation techniques} are introduced later in the paper\footnote{\textcolor{red}{The best 
anonymization technique is the one that reduces the accuracy 
of sensitive inferences to the level of random guessing. As we discuss later, 
we can get close to this by manipulating latent representations in a probabilistic fashion.}}.
The modified latent representation is then sent to the decoder 
to construct the anonymized/obfuscated version of the original input data embedding.
It is worth mentioning that apart from the central server, we do not trust any other party. 
Even the central server does not know about public and private attributes of the data 
that is being obfuscated at the edge. 
It simply sends all average representations to edge devices and they select the appropriate ones 
based on the predicated attributes of their input data.}


We evaluate our anonymization technique on two publicly available Human Activity Recognition (HAR) datasets, 
namely MotionSense~\cite{malekzadeh19} and MobiAct~\cite{vavoulas2016mobiact}. 
We use MotionSense in a two-class gender anonymization task. 
The MobiAct dataset is used in two additional anonymization tasks, 
namely two-class gender anonymization and multi-class weight anonymization. 
The contribution of this paper is threefold:
\begin{itemize}
    \item We propose a VAE-based anonymization technique 
    that learns and manipulates latent representations of the input sensor data 
    to effectively prevent sensitive inferences on the obfuscated data.
    We modify the loss function of the VAE by incorporating a private attribute classification loss term. 
    This term helps the VAE learn more useful and anonymization-friendly representations.
    We further train a different VAE model for each public attribute and 
    show that it can effectively reduce the size of the model. 
    This is crucial for performing anonymization at the edge. 
    
    \item We compare the performance of our technique with autoencoder-based anonymization techniques proposed in~\cite{malekzadeh19, malekzadeh2017replacement} \textcolor{red}{and the general VAE-based anonymization technique introduced in~\cite{hajihass2020latent}} by evaluating it on MotionSense and MobiAct datasets.
    \textcolor{red}{We show that our anonymization technique can reduce the accuracy of sensitive inferences 
    by up to $80\%$ and is less vulnerable to the private attribute re-identification attacks. 
    In particular, it reduces the leakage of information about private attributes
    by 23\% more that AAE~\cite{malekzadeh19} on average in the three anonymization tasks.}
    
    \item We evaluate the feasibility of performing adversarial model-free anonymization in real time
    on edge devices by running experiments on a Raspberry Pi 3 Model B. 
\end{itemize}

The rest of this paper is organized as follows. 
Section~\ref{related} summaries related work 
on privacy-preserving data analysis and data anonymization. 
\textcolor{red}{Section~\ref{threat} presents the threat model and 
a specific type of attribute inference attack concerning the private attribute, 
which we refer to as the \textcolor{red}{re-identification} attack.}
Section~\ref{back} provides the necessary background on VAE, 
introduces the idea of adversarial model-free anonymization,
and explains the distinction between adversarial model-based and model-free anonymization techniques.
Section~\ref{proposed} describes our adversarial model-free anonymization technique, 
the modified VAE loss function, and the specialized VAEs.
The datasets and evaluation results of the proposed anonymization technique are presented in Section~\ref{eval}. 
Furthermore, we evaluate our technique with both deterministic and probabilistic modifications
of the latent space representation and 
investigate whether they can successfully prevent the re-identification attack.
In Section~\ref{implementation}, we study real-time anonymization of sensor data on an edge device. 
Section~\ref{conc} concludes the paper and provides directions for future work.
\section{Related Literature}
\label{related}
Most IoT devices today are equipped with a myriad of sensors that collect data from 
people and their surrounding environment. 
The sheer amount of personally identifiable and sensitive information embedded in this time series data
opens the door for unwanted and private attribute inferences.
In~\cite{ren2019information}, information exposure from 81~consumer IoT devices 
is analyzed with respect to their network traffic. 
It is found that 72 out of the 81~IoT devices send data to a third-party over the Internet. 
This underscores the importance of anonymizing data before it leaves the IoT device.

The literature on data anonymization and privacy-preserving data analysis is extensive. 
Related work can be broadly classified into systemic and algorithmic solutions. 
The systemic solutions provide mechanisms for monitoring and managing access to private or sensitive data,
and efficiently masking or downsampling this data~\cite{Gotz2012, Chakraborty14, singh2018tussleos}. 
The algorithmic solutions can be further divided into 
solutions based on \emph{differential privacy} and \emph{k-anonymity}~\cite{dwork2011differential, Comas2014, bayardo2005data},
and solutions that rely on \emph{deep generative models}~\cite{malekzadeh19, malekzadeh2017replacement, feutry2018learning, hajihass2020latent}.
The former category hides an individual's private data in a population,
whereas the latter obfuscates data in a certain way to limit the risk of disclosing private attributes.
These techniques have been applied in a variety of domains, 
including public health~\cite{Dankar2012, phan2016differential}, and
smart homes and buildings~\cite{wu2019privacy, brkic2017know, jia2017privacy, jia2017pad}.


\textbf{Systemic solutions.} There are various approaches to enhancing data privacy at
the operating system and firmware level~\cite{fernandes2016flowfence, singh2018tussleos}, 
at the application level~\cite{mo2020darknetz,osia2020hybrid}, and via certain protocols~\cite{carbunar2010query}. 
The operating system solutions enable the user to
navigate the tradeoff between privacy and functionality of IoT devices~\cite{singh2018tussleos, fernandes2016flowfence}. 
Reference~\cite{fernandes2016flowfence} promotes user privacy through sand boxed execution 
of developers' code in quarantined execution modules and taint-tracked data handlers. 
Taint-tracked opaque data handlers help prevent applications from accessing sensitive data
and sharing such data via the network interface. 
In~\cite{singh2018tussleos}, a privacy abstraction technique is proposed 
to manage control over sensor data tussles, addressing utility-privacy tradeoffs. 
Other architectural solutions try to avoid data and model information leakage 
in data processing at the edge and in the cloud. 
In~\cite{carbunar2010query}, 
an efficient privacy-preserving querying protocol is proposed for sensor networks
assuming that client queries are processed by servers controlled by multiple mutually distrusting parties. 
These queries reveal both an array of specific sensors and relationships between the subsequent queries. 
This will dissuade the organizations from sharing resources to build large-scale shared sensor networks. 
To address these risks, the authors propose the SPYC protocol which guarantees privacy protection
if servers do not cooperate in attacking the clients. 
They also discuss possible solutions when servers cooperate to infringe privacy of clients. 
These systemic solutions are designed for specific use cases and do not address the data anonymization problem in general.

In Federated Learning~\cite{konevcny2016federated}, 
clients, which can be edge devices, do not share their data with a central server to train a model.
The model is instead trained in a number of iterations by the clients using their own data
and the updated model is sent to the server after each iteration.
While this eliminates the need to share private user data with a third-party,
it will not protect private user data from on-device inferences that are intrusive.

\textbf{Algorithmic solutions.} 
Algorithmic solutions are based on microaggregation and various machine learning techniques. 
The techniques that are based on $k$-anonymity and differential-privacy 
include~\cite{dwork2008differential, Comas2014, bayardo2005data, erlingsson2014rappor}. 
The authors in~\cite{jia2017privacy} propose a privacy aware HVAC control system architecture 
that decreases the privacy risk subject to some control performance guarantee 
using the mutual information (MI) metric.
PAD is a privacy preserving sensor data publishing framework which is proposed in~\cite{jia2017pad}.
It ensures privacy through data perturbation
and works with customized user datasets with configurable privacy constraints for end users. 
The authors in \cite{sangogboye2018framework} propose enhancements to PAD, 
allowing it to be used with non-linear features. 
In~\cite{he2011pda}, authors propose two different techniques for private data aggregations namely, Cluster-based Private Data Aggregation (CPDA) and Slice-Mix-AggRegaTe (SMART) based on different properties. 
These techniques are used to provide efficient data aggregation while protecting the user's data privacy. 
Furthermore, there exists cryptographic privacy-preserving techniques, such as~\cite{miao2019privacy}, 
which proposes weighted aggregation on users' encrypted data 
using a homomorphic cryptosystem to promote 
privacy-preserving crowd sensing systems in the truth discovery domain.
This ensures both high privacy protection and high data utility.

Other algorithmic solutions rely on machine learning models, 
from deep neural networks (DNN) to generative models, 
such as generative adversarial models (GAN)~\cite{goodfellow2014generative} and VAE~\cite{kingma2013auto}.
DNNs are used to protect the privacy of patients by de-identifying patients' personal notes 
through the removal of personal health identifiers~\cite{dernoncourt2017identification}. 
For security devices, such as home security cameras and baby monitors,
research has focused on de-identifying personal attributes from the camera feed. 
These approaches are essential due to the lack of trust 
between users and cloud providers~\cite{wu2019privacy, brkic2017know}. 
Many security devices use cloud servers for data storage and processing. 
Contrary to crude facial blaring or pixelization techniques 
that have been adopted in the past (i.e., adding noise and masking), 
GANs can be used to swap faces in a camera feed~\cite{wu2019privacy}. 
Thus, less information is lost as the facial expressions can be kept intact 
while the user identity is protected. 
\textcolor{red}{Another technique called PrivacyNet~\cite{mirjalili2020privacynet} uses 
a GAN-based semi-adversarial network to change an input face image 
such that it can be used for biometrics purposes but not for reliable identification of attributes.
PrivacyNet allows for choosing which specific attributes must be obfuscated in the input face images (e.g., age and gender) and which are be extracted (e.g., ethnicity).
These papers show the usefulness of machine learning models to address privacy issues 
in imagery and video data, but they do not address the privacy-utility tradeoff 
for time series data gathered by IoT devices.}

\textcolor{red}{There is also a variety of machine learning-based techniques 
that address the utility-privacy tradeoff; Table~\ref{tab:related} provides a summary of these papers.
References~\cite{malekzadeh19, huang2018generative, liu2019privacy, Olympus2019} 
use autoencoders trained in an adversarial fashion to anonymize sensor data.
We refer to them as \emph{adversarial model-based} anonymization techniques.
Specifically, the private attribute (either categorical or binary) is obscured 
by solving a minimax optimization problem that involves an adversary and an obfuscator.
The Privacy Adversarial Network (PAN)~\cite{liu2019privacy} generates 
task-specific features that do not disclose the private attribute. 
Should this technique be employed for data anonymization, 
existing applications that rely on raw sensor data cannot be used.
In~\cite{huang2018generative}, it is assumed that some statistics of the dataset
are known to the obfuscator. This assumption does not always hold in practice.} 

\textcolor{red}{The closest lines of work to ours are~\cite{malekzadeh19,Olympus2019}.
We argue that due to the reliance of these techniques on adversarial training, 
they merely fool the adversarial model used during training and 
remain vulnerable to the re-identification attack (defined in the next section).
We compare our technique with~\cite{malekzadeh19} to show that their technique falls prey to this attack.
Olympus~\cite{Olympus2019} tries to address this problem by using a hypothesis space of models, 
in particular DNNs, for adversarial training.
While this provides a significant improvement over the other adversarial model-based techniques, 
it still does not work quite well when the attacker uses a different class of models (e.g., random forests) 
than the models used for adversarial training.
In this work, we propose the use of an adversarial model-free anonymization technique 
that randomly modifies the latent representation of the input data 
to obfuscate the data and at the same time prevent the re-identification attack to some extend.}

We extend our previous work~\cite{hajihass2020latent} by improving the anonymization capability of 
our adversarial model-free technique and studying the feasibility of performing anonymization on edge devices. 
Moreover, we investigate whether the proposed technique can successfully transform latent space representations
and reconstruct data when the private attribute class is not binary (i.e., has multiple classes).

\begin{table}[]
\caption{\textcolor{red}{Related work on anonymizing sensor data via generative models}}
\resizebox{\columnwidth}{!}{%
\begin{tabular}{|c|l|c|l|}
\hline
Reference & \multicolumn{1}{c|}{Anonymization technique} & Adversarial training? & \multicolumn{1}{c|}{Brief description} \\ \hline
\cite{malekzadeh19} & Anonymizing Autoencoder (AAE) & \checkmark & \begin{tabular}[c]{@{}l@{}}
An autoencoder is trained to remove personal information from sensor data. This work uses a specific adversarial\\
model to train the autoencoder. We use this work as a baseline.
\end{tabular} \\ \hline
\cite{malekzadeh2017replacement} & Replacement Autoencoder (RAE) & {} & \begin{tabular}[c]{@{}l@{}}
An autoencoder that replaces sensitive sections of time series with non-sensitive sections. It assumes these\\
sections are non-overlapping and can be separated by the user. We do not make these assumptions in this work.  
\end{tabular} \\ \hline
\cite{huang2018generative} & Generative Adversarial Privacy (GAP) & \checkmark & \begin{tabular}[c]{@{}l@{}}
GAP formulates a minimax game between the data owner and adversary to anonymize private attributes. It assumes\\ that private attributes are included in the dataset and the data owner knows some statistics of the datasets.\\ Unlike this work, we assume that private attributes are not part of the dataset, but can be inferred from\\
time series data. We also do not make assumptions about the distribution of private and public attributes.
\end{tabular} \\ \hline
\cite{liu2019privacy} & Privacy Adversarial Network (PAN) & \checkmark & \begin{tabular}[c]{@{}l@{}}
An encoder is adversarially trained to learn representations that only convey information about the target task.\\
It differs from our work because it learns features that do not contain information about the private attribute,\\
whereas we attempt to reconstruct the sensor data.
\end{tabular} \\ \hline
\cite{Olympus2019} & Utility Aware Obfuscation (Olympus) & \checkmark & \begin{tabular}[c]{@{}l@{}}
An obfuscation mechanism is learned using an adversarial game such that the privacy loss and the accuracy loss\\
are jointly minimized. This work adopts adversarial networks to model privacy and utility requirements.
\end{tabular} \\ \hline
\cite{hajihass2020latent} & VAE-based Mean Manipulation & {} & \begin{tabular}[c]{@{}l@{}}
A VAE with the original ELBO is trained for anonymization through mean manipulation. This is our previous work\\
which we extend by augmenting the loss function with the cross-entropy loss of a classification layer.
\end{tabular} \\ \hline
\end{tabular}
}
\label{tab:related}
\end{table}
\section{Threat Model}\label{threat}

\textcolor{red}{
\noindent\textbf{User Data.} 
We focus on the time series data generated by one or multiple sensors in the IoT device.
This data can be stored and processed locally or sent to the cloud for aggregation and further analysis.
We assume there are patterns in this data that reveal public and private attributes of the user.
These attributes can take a limited number of values and we assume the set of possible values 
they can take is known in advance. 
We distinguish between the two types of attributes based on whether 
inferring them from raw sensor data is desirable for the user.
Private attributes often include sensitive information about a user (e.g., sex, age, weight, race) 
and their environment (e.g., size, occupancy).
The user needs to specify the private attribute they would like to obscure 
and the public attribute they would like to remain intact 
in order to use the obfuscator which is introduced next.}

\textcolor{red}{\noindent\textbf{Applications.} There are typically several third-party applications 
that have access to user data in the edge device or in the cloud.
Although these applications cannot be trusted, they offer benefits that are important to the user.
In most cases, they operate on the sensor data and may have 
internal models to predict the public attribute and use it in different ways.
Our goal is to preserve the functionality of these applications by obfuscating sensor data
rather than predicting the public attribute using our own classification model 
and releasing it instead of the sensor data.
}

\textcolor{red}{
\noindent\textbf{Obfuscator.} The obfuscator is the anonymization networks we propose in this work;
it runs on the IoT/edge device to obfuscate sensor data such that the risk of disclosing 
the private user attribute is limited, while the usefulness of this data for applications and services that 
rely on raw sensor data to infer public user attributes is not changed significantly.
We assume sensor data always goes through the obfuscator 
before it can be used by applications\footnote{\textcolor{red}{This may require operating system level support, which has been studied in the literature but is outside the scope of this work.}} or sent to the cloud.}

\textcolor{red}{
\noindent\textbf{Attacker.} The attacker can be any party with interests in 
identifying the user's private attribute by analyzing their sensor data.
The attacker is able to secure access to raw sensor data through malicious applications 
that run on the edge device or in a cloud server provided that sensor data is transferred to the cloud.
The attacker can be a data broker, an advertising agency, a social media company, or a government.}

\textcolor{red}{
\noindent\textbf{Re-identification Attack.} It is a type of property inference attack in which 
the attacker aims to re-identify a private attribute in the obfuscated sensor data.
To perform this attack, the attacker first needs to obtain enough labelled sensor data 
where the label represents the actual private attribute.
Next, they pass this data through the obfuscator and record the output, 
which is the obfuscated version of this data.
Finally, they train a model that maps the obfuscated data to the actual private attribute.
Once trained this model can be used to de-anonymize the sensor data and disclose the private attribute.}

\begin{figure}[t]
\centering
  \includegraphics[scale=.6]{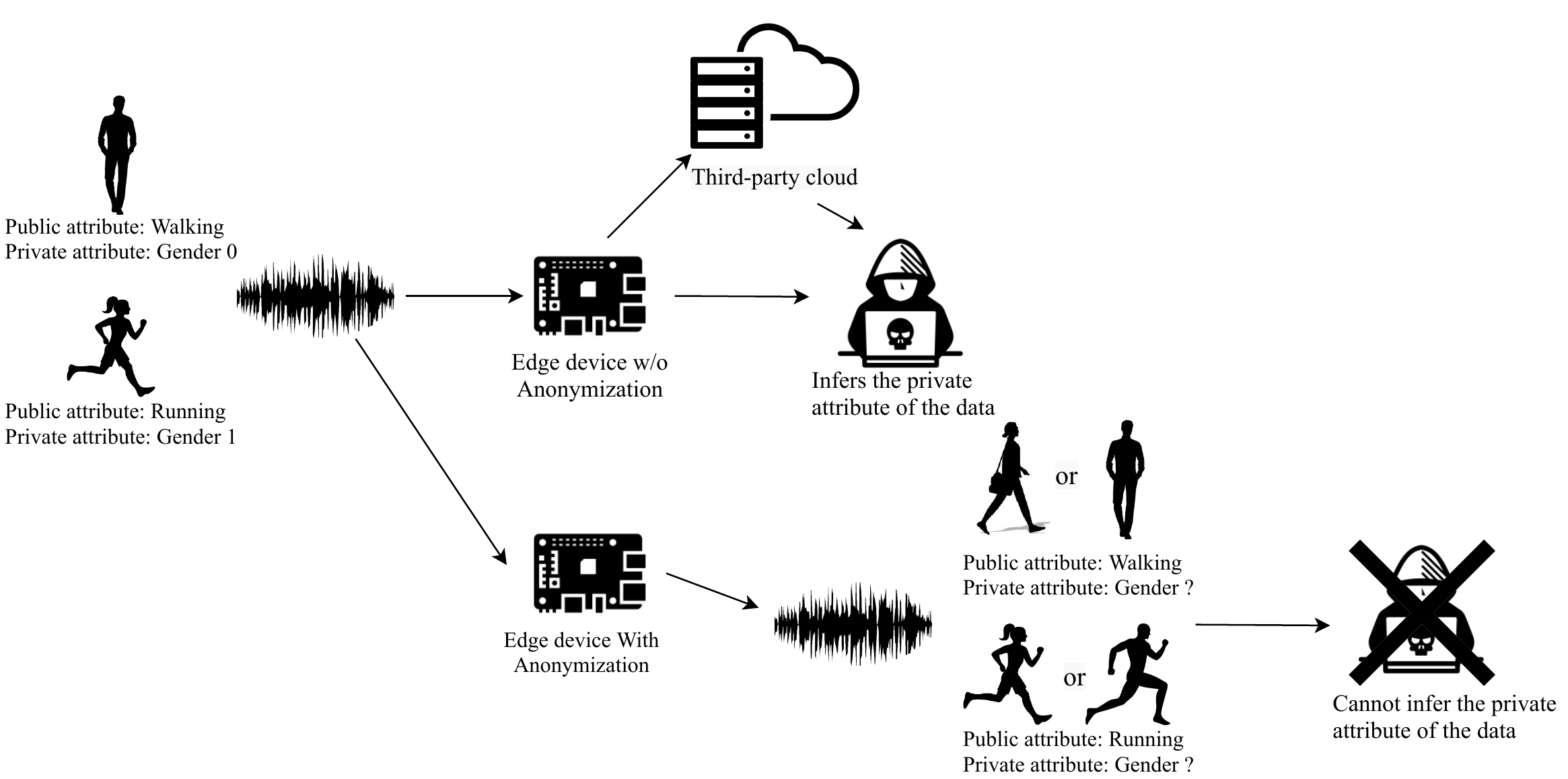}
  \caption{\textcolor{red}{An example scenario in which a fitness tracker user 
  would like to infer their activity while preventing the attacker from inferring their gender.
  To this end, they run the proposed anonymization technique on the edge device to obfuscate the sensor data
  before it is used by an untrusted activity recognition application. 
  This application may run on the edge device or in the cloud as shown here.}}
  \label{fig:threatmodel}
\end{figure}

\textcolor{red}{
\noindent\textbf{Example Scenario.} To better describe the threat model,
we consider an example scenario depicted in Figure~\ref{fig:threatmodel}.
In this scenario, data is generated by the inertial measurement unit 
embedded in a fitness tracker and is stored in an IoT/edge device.
The user wishes to infer the activity type, which is the public attribute, 
from the sensor data to calculate the duration of each activity.
There are several applications that work on the raw sensor data and can be used for this purpose.
Some of them can be installed on the edge device to make the desired inferences locally,
while others are cloud services that require transferring sensor data to a cloud server.}

\textcolor{red}{
Unfortunately, these third-party applications cannot be trusted 
as they may also infer private user attributes, 
such as their gender, from this data without the user's consent.
In this case, the application developer is the attacker.
As we discuss in the next section, obfuscation is done such that the user activity remains intact, 
while the gender identity of the user is changed either at all times or at random.}

\section{Background}
\label{back}
\subsection{Variational Autoencoders}
\begin{figure}[h]
\centering
  \includegraphics[scale=0.7]{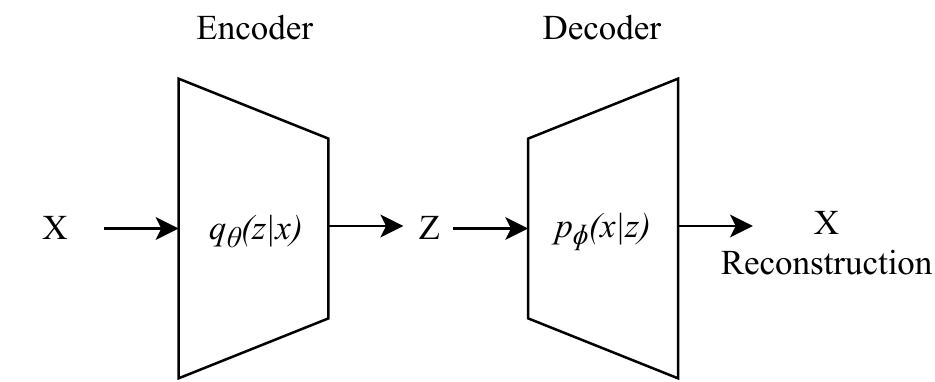}
  \caption{Probabilistic encoder and decoder networks of a VAE.}
  \label{fig:vae3}
\end{figure}

A Variational Autoencoder (VAE)~\cite{kingma2013auto} is a generative model 
comprised of an encoder and a decoder network. 
It differs from a standard autoencoder as it models an underlying probability distribution 
over the latent variables which is quite useful for the synthesis process, 
hence we use it in this work\footnote{
We also found empirically that modifying latent representations of a VAE 
(following the procedure outlined in Section~\ref{proposed}) 
can more effectively confuse the intrusive inference model and reduce the privacy loss 
than modifying latent representations of a vanilla autoencoder.}.
Figure~\ref{fig:vae3} shows the probabilistic encoder and decoder of a VAE.
The probabilistic encoder represents the approximate posterior 
in the form of $q_{\theta}(z|x)$ (e.g., a multivariate Gaussian with diagonal covariance). 
After training the network parameters denoted by $\theta$, $q_{\theta}(z|x)$ is used 
to \emph{sample} a latent space representation, $z$, for a given data point, $x$.
The data reconstruction is governed by the likelihood distribution $p_{\phi}(x|z)$ 
which is modeled by the probabilistic decoder.

We minimize a loss function to train a variational autoencoder.
The objective here is to increase the quality of the reconstruction
performed by the decoder (i.e., maximizing the log-likelihood of $p_{\phi}(x|z)$), 
while having the encoder learn meaningful and concise representations of the input data 
(minimizing the Kullback-Leibler (KL)-divergence between the approximate and true posterior).
Concretely, the objective is to find the best set of parameters (i.e., weights and biases) 
in the probabilistic decoder and encoder that minimize the reconstruction error of the input data 
given the latent variables while approximating a variational posterior distribution $q_{\theta}(z|x)$ 
that resembles the true posterior distribution $p(z|x)$. 

Since the distance between the variational and true posterior cannot be calculated exactly, 
the Evidence Lower Bound (ELBO) of a variational autoencoder is maximized 
to minimize the KL-divergence term between the approximated variational posterior and 
the prior over latent variables, $p_{\phi}(z)$, which is assumed to have a Gaussian probability distribution.
The ELBO is the negative of the loss function and can be written as:
\begin{equation}\label{eq:ELBO}
\text{ELBO}_i(\phi, \theta) = \mathbb{E}_{z\mathtt{\sim}q_{\theta}(z|x_i)}\log{p_{\phi}(x_i|z)}-\text{KL}\big(q_{\theta}(z|x_i)||p_{\phi}(z)\big)    
\end{equation}
This lower bound is calculated above for a single data point $x_i$,
hence the loss function is obtained by summing this over all data points in the dataset. 
The KL-divergence can be viewed as a regularizer and a constraint on the approximate posterior.
The ELBO with the Gaussian assumption (with mean $\mu$ and standard deviation $\sigma$) 
for the latent and the approximated posterior distribution is given below:
\[\mathbb{E}_{z\mathtt{\sim}q_{\theta}(z|x_i)}[\text{log}\: p_{\phi}(x_i|z)]+\frac{1}{2}\sum_{j=1}^J [1+\text{log}\:(\sigma^{(j)}_i)^2-(\sigma^{(j)}_i)^2-(\mu^{(j)}_i)^2]\]
where $j$ is the index of a latent variable in the latent representation $z\in\mathbb{R}^J$.

Recent work on the disentanglement of the latent representations 
shows that penalizing the KL-divergence term can help to achieve better disentanglement of latent variables. 
Concretely, a higher weight should be assigned to the KL-divergence term
for latent variables to represent distinct features of the data. 
The weight factor is denoted by $\beta$ in the following
\[\text{ELBO}_i(\phi, \theta) = \mathbb{E}_{z\mathtt{\sim}q_{\theta}(z|x_i)}\log{
p_{\phi}(x_i|z)}-\beta\;\text{KL}\big(q_{\theta}(z|x_i)||p(z)\big)\]
The $\beta$ value in the original VAE is equal to $1$ which would be multiplied by $\frac{1}{2}$ 
given the Gaussian assumption.
It is argued in~\cite{higgins2017beta} that by choosing a $\beta>1$ in the ELBO, 
more disentangled latent representations can be learned. 
However, higher disentanglement degrades the reconstruction accuracy.
This highlights the intricate tradeoff in the training of VAEs.

\subsection{Adversarial Model-based versus Model-free Anonymization}
We refer to the models used for making desired and unwanted inferences 
as the public and private attribute inference models, respectively.
The output of the public attribute inference model is either a single public attribute class or a distribution over all public attribute classes.
Similarly, the output of the private attribute inference model is either a single private attribute class 
or a distribution over all private attribute classes.
For example, in a fitness-tracking application, the public attribute inference model outputs an activity label
while the private attribute inference model could assign the user to a particular age group.
We assume that this application is not supposed to learn the age of the user.

To protect private and sensitive data against a wide range of private inference models 
rather than a specific model, we propose adversarial model-free anonymization.
Unlike adversarial model-based anonymization techniques 
which use a specific private inference model in adversarial training,
adversarial model-free anonymization techniques utilize the definition of the unwanted inference
to apply a linear transformation to latent space representations. 
This transformation should have imperceptible impact on data utility and support privacy-preserving inferences.

\textcolor{red}{The adversarial model-free anonymization technique which we proposed in this work 
utilizes a deep generative model (i.e., a VAE) to learn and subsequently manipulate 
the latent representation of input data.
This is one advantage of our adversarial model-free anonymization technique 
over its adversarial model-based counterparts.
By modifying the latent representation, we ensure that the reconstructed data 
has a different private attribute class label than the original data. 
This modification can be either deterministic or probabilistic. 
The randomness introduced when the modification is done with some probability, 
makes our work less susceptible to the re-identification attack.
We dive into detail about latent space transformations in the next section.}


\section{Methodology}
\label{proposed}
In this section, we describe how we learn a useful representation
for an embedding of time series data (generated by a sensor)
using a VAE with a modified loss function.
The choice of which attribute-specific VAE to use depends on the predicted public attribute class (i.e., the output of the public attribute inference model).
We explain how this representation can be transformed 
given the predicted public attribute class and the predicted private attribute.
These steps are illustrated in Figure~\ref{fig:modvae}.

\subsection{A Modified Loss Function for VAE}
The loss function we use in this paper 
builds on the original VAE's loss function proposed in~\cite{kingma2013auto}. 
We modify this loss function by adding an extra term that corresponds to the classification error 
of the private attribute classifier, i.e., $Enc_\theta + f_\eta$. 
Specifically, the encoder network is supplemented with a classification layer, $f_\eta$, 
to encourage learning representations that are more representative 
of the private attribute associated with the input data. 
\textcolor{red}{This differs from adversarial training 
because the classification layer is not trained to fool the network.
Instead it works in tandem with the VAE to ensure the learned representations 
are more separable along the private attribute.}
We first introduce this loss function and then discuss why minimizing 
this function can result in a more effective anonymization. 
The augmented loss function can be written as:
\begin{equation}\label{eq:modloss}
    -\sum_{k=1}^{K}{ \bigg(\mathbb{E}_{z_k\mathtt{\sim}q_{\theta}(z_k|x_k)}
\left[\log p_{\phi}(x_k|z_k)\right] - \beta\;D_{\text{KL}}\big(q_{\theta}(z_k|x_k)||p(z_k)\big)}
+\alpha\;\sum_{i=1}^{M}{y^i_k \log\big(f^i_\eta(z_k)\big) \bigg)}
\end{equation}
where $z_k$ denotes the latent representation of the $k_{th}$ input data embedding,
$y_k$ denotes the true private attribute class label of that embedding\footnote{Note that $y^i_k$ is $1$ 
if and only if $x_k$ belongs to the private attribute class $i$, and is $0$ otherwise.},
and $f_\eta$ is the classification layer. 
\textcolor{red}{
We assume that the $K$ input entries are sampled from a discrete or continuous distribution.
$p_\phi(x_k|z_k)$ is referred to as the probabilistic decoder portion of the VAE model 
which generates distributions over possible $x_k$ values given $z_k$.
Likewise, $q_\theta(z_k|x_k)$ is referred to as the probabilistic encoder 
which produces distributions over possible $z_k$ latent representations given $x_k$. 
The learned distribution over latent representations given $x_k$ 
can be a multivariate Gaussian or a Bernoulli distribution. 
In our case, we choose a multivariate Gaussian since we are dealing with real-valued data.
Note that the first two terms in this loss function are the two terms in Equation~(\ref{eq:ELBO}). 
The only difference is the introduction of the $\beta$ weight factor 
for the Kullback–Leibler divergence term as explained in~\cite{higgins2017beta}.} 

The main limitation of the $\beta$-VAE's loss function for data anonymization
is the inherent tradeoff between the quality of the reconstructed data and 
the disentanglement of the learned latent representations. 
In general, lower $\beta$ values would yield better accuracy in the data reconstruction task 
(higher data utility),
and higher $\beta$ values would train the VAE to generate more disentangled latent representations 
(lower private attribute inference model accuracy).

The best anonymization performance by a VAE is achieved when the data utility is the highest and 
the accuracy of the private inference model is the lowest.
Thus, we need to tweak the loss function to have the highest data utility in the anonymized data 
(determined by the reconstruction loss and KL-divergence),
while having as much disentanglement as possible (determined by KL-divergence) 
for the lowest private inference accuracy.
As discussed in~\cite{higgins2017beta},
there is a limit to the learning capacity of a conventional VAE's loss function.
Hence, to increase the anonymization capability of the trained VAE, 
we add the private-attribute classification loss to the ELBO.
Specifically, we use the latent representation of the original input data 
as input to a single-layer neural network which infers the private attribute class of each data embedding. 
This neural network, represented as $f_\eta$, will be trained alongside the VAEs encoder. 
In essence, the classification layer, $f_\eta$, and the VAEs encoder together form a classification network. 
The learned latent variables become more representative of private attributes in the data consequently.

\begin{figure}[t!]
\centering
  \includegraphics[scale=.62]{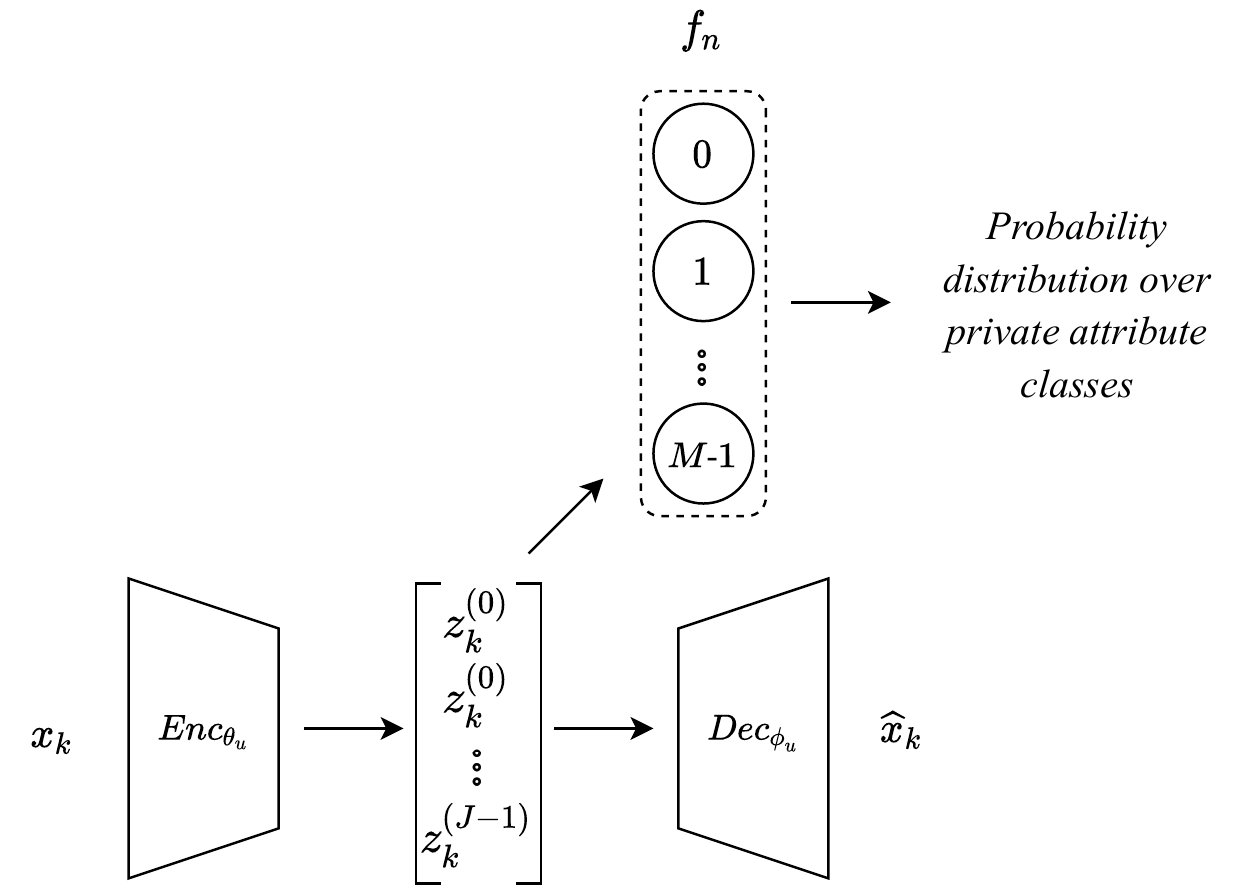}
  \caption{Variational Autoencoder with an additional classification layer denoted by $f_{\eta}$.}
  \label{fig:class}
\end{figure}

We use the cross entropy loss which is the distance between the predicted private attribute class of 
each anonymized data embedding and its ground truth value, $y^i_k$.
We create a simple classification layer that maps the latent representations generated by VAE 
to the private attribute class labels of each of the corresponding input data entries as illustrated in Figure~\ref{fig:class}. 
Thus, the addition of the classification loss to the loss function 
encourages the VAE to learn more anonymization-friendly representations.

We argue that adding the classification layer, $f_\eta$, will force the probabilistic encoder 
to learn latent representations that are separable along the private attribute class labels, $y_k$'s. 
Our results confirm that the added term to the objective function improves the performance of VAE 
in the anonymization task by introducing some structure and enforcing a clear separation 
between different classes in the latent space.
We instantiate $f_\eta$ as a single layer of neurons with a softmax activation function.
In particular, this layer contains $M$ neurons, $n_0, ..., n_{M-1}$,
where $M$ represents the number of private attribute classes in the original dataset. 
The trainable set of weights used by the classification layer is denoted by $\eta$. 
Suppose each latent representation is a vector of $J$ latent variables, $z^{(0)}_k, \cdots, z^{(J-1)}_{k}$. 
Thus, each $\eta_j^m$ is the weight connecting input $z^j_k$ to neuron $n_m$. 
The output of the $m^{th}$ neuron in the classification layer can be written as
$z_k \eta_m^\top$. The output of all the neurons goes through softmax activation 
to produce a probability distribution over private attribute classes given the input data:
$f^m_{\eta}(z_k) = \frac{e^{z_k \eta_m^\top}}{\sum_{i}e^{z_k \eta_i^\top}}$.


The two hyperparameters in Equation~(\ref{eq:modloss}), namely $\alpha$ and $\beta$, 
must be tuned for each VAE as discussed later.
The VAE and the classification layer are depicted in Figure~\ref{fig:class}. 


\subsection{Representation Learning with a VAE Customized for Each Public Attribute Class} 

We train a VAE for each public attribute class in our dataset. 
By having attribute-specific VAEs instead of just one general VAE, 
which learns latent representations for all input data 
regardless of their public and private attribute classes~\cite{hajihass2020latent},
we break down the model into multiple models that are smaller in size. 
Each of these models is trained to reconstruct data for a given public attribute class.

One key advantage of using public attribute-specific VAEs is the reduction in the size of the model. 
It also allows for applying a higher disentanglement constraint 
(i.e., the weight $\beta$) in the training process. 
We get a $12$-fold reduction in the model size and the number of trainable weights, 
from roughly $24$ million weights in the case of a general VAE 
to a total of $2$ million weights for all attribute-specific VAEs. 
Moreover, using parsimonious models enhances the anonymization performance 
when compared to~\cite{hajihass2020latent}.

Since we have multiple public attribute-specific VAE models, 
it is necessary to predict the public and private attribute classes of a given data embedding 
at anonymization time.
This information is used to determine which VAE must be selected for anonymization.
We do this using the pretrained classifiers shown in Figure~\ref{fig:modvae}.

\subsection{Transforming Latent Representations}
Algorithm~\ref{algorithm:1} shows different steps of the proposed 
anonymization technique (\textcolor{red}{labelled 1 to 6})
assuming that a VAE is trained already for each public attribute class.
This algorithm operates on fixed-size embeddings of the input time series data.
These embeddings are created by considering a window 
that contains a number of consecutive data points in the time series.
After the first embedding, a new embedding is created after a certain number of new data points are received (determined by the stride length).

Suppose we have $k$ data embeddings denoted by $x_1, \cdots, x_k$.
Each embedding has corresponding public and private attributes.
The proposed anonymization algorithm takes as input an embedding along with 
encoder and decoder parameters of different VAEs, and
the average latent representation denoted by $\bar{z}^{i}_{u}$ 
for each public attribute class $u$ and private attribute class $i$.
These average representations are calculated for the training dataset in the cloud 
or at the edge provided that the edge device retains a copy of the whole training dataset.

In \textcolor{red}{Step~1 of the algorithm}, the pretrained public and private attribute classifiers 
(not to be confused with the classification layer $f_\eta$) 
are used to identify the public attribute class $u$ and 
the private attribute class $i$ for each data embedding $x_k$.
After inferring the public and private attribute classes, 
we load $\text{Enc}_{{\theta_u}}$ and $\text{Dec}_{{\phi_u}}$ models 
for the inferred public attribute class $u$. 
\textcolor{red}{In Step~2,} the encoder part of this attribute-specific VAE encodes $x_k$ in a probabilistic manner. 
The corresponding latent representation, $z_k$, can be sampled from the distribution.
Once the representation is sampled, 
we change the inferred private attribute class label of $x_k$ via a simple function 
which we refer to as \texttt{Modify}\textcolor{red}{, this is shown in Step~3.}
This function converts the inferred private attribute class label of $x_k$ from $i$ 
to an arbitrary private attribute class label denoted by $i^{'}$.

\begin{algorithm}[t]
\SetAlgoLined
\KwData{data embedding $x_k$, 
average latent representations, autoencoder parameters $\theta_u~\text{and}~\phi_u$ 
for each public attribute class $u$,
pretrained classifiers for public and private attributes}
\KwResult{anonymized data embedding $\hat{x}_{k}$}
    \textcolor{red}{\textbf{1}} \quad $u, i$ $\gets$ $\text{Classify}$($x_k$) \;
    
    \textcolor{red}{\textbf{2}} \quad $z_k$ $\gets$ $\text{Enc}_{{\theta_u}}(x_{k})$ \; 
    
    \textcolor{red}{\textbf{3}} \quad $i^{'}$ $\gets$ $\text{Modify}(i)$ \textcolor{red}{\tcp*{deterministic or probabilistic}}
    
    \textcolor{red}{\textbf{4}} \quad ${\bar{z}}^{i}_{u}, {\bar{z}}^{i'}_{u} \gets \text{Load mean latent representations}$ \textcolor{red}{\tcp*{from mean representations sent by central server}}
    
    \textcolor{red}{\textbf{5}} \quad $\hat{z}_{k}$ = $z_{k}-{\bar{z}}^{i}_{u} + {\bar{z}}^{i'}_{u}$ \;
    
    \textcolor{red}{\textbf{6}} \quad $\hat{x}_{k}$ $\gets$ $\text{Dec}_{{\phi_u}}(\hat{z}_{k})$ \;
 \caption{Adversarial model-free anonymization with representation learning and transformation}\label{algorithm:1}
\end{algorithm}

\begin{figure}[t]
\centering
  \includegraphics[scale=.8]{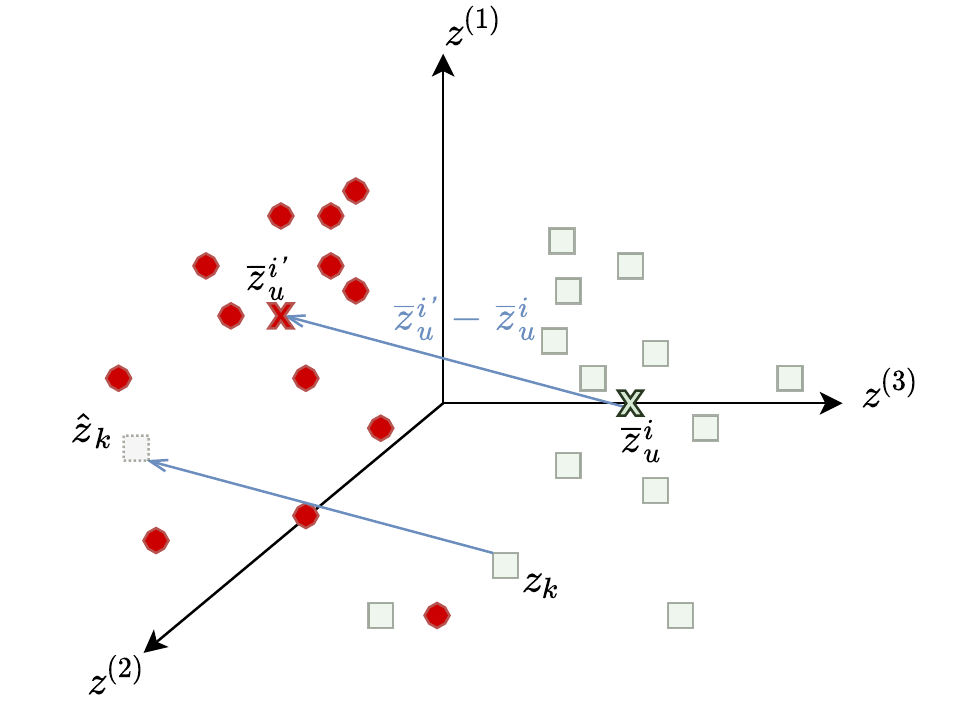}
  \caption{Overview of the adversarial model-free anonymization technique assuming a 3-dimensional latent space.
  Latent representations which have a public attribute other than $u$ are not shown in this figure. \textcolor{red}{In this figure, $z^{(1)}, z^{(2)}$, and $z^{(3)}$ are the three dimensions of our exemplary latent space. $z_k$ and $\hat{z}_k$ are the corresponding latent representation and modified latent representation of the $k_{th}$ input data embedding. $\bar{z}^i_u$ represents the average value of all latent representations pertaining to the public attribute $u$ and private attribute $i$. Moreover, ${\bar{z}}^{i'}_{u}-{\bar{z}}^{i}_{u}$ is the transfer vector.}}
  \label{fig:transformation}
\end{figure}

The transformation of a latent representation involves a sequence of simple arithmetic operations. 
Consider a representation $z_k$ with public attribute $u$ and private attribute $i$,
and let us denote the average of all latent representations with public attribute $u$ and private attribute $i$ 
by ${\bar{z}}^{i}_{u}$.
\textcolor{red}{In Step~4, we select the two average latent representations, ${\bar{z}}^{i}_{u}, {\bar{z}}^{i'}_{u}$, among the average latent representations sent by the central server.
The central server is responsible for calculating these average representations 
using a representative dataset and communicating them to the edge devices.}
\textcolor{red}{In Step~5,} we obtain the transformed representation of $x_k$, denoted $\hat{z}_{k}$,
by subtracting ${\bar{z}}^{i}_{u}$ from $z_k$ and 
adding ${\bar{z}}^{i^{'}}_{u}$ to the result.
The probabilistic decoder, $\text{Dec}_{{\phi_u}}$, 
takes $\hat{z}_{k}$ instead of $z_k$ to synthesize data. 
We use the term \emph{transfer vector} to refer to ${\bar{z}}^{i'}_{u}-{\bar{z}}^{i}_{u}$, 
which is the Euclidean distance between 
the average of all representations with private attribute $i$ and public attribute $u$ and
the average of all representations with private attribute $i^{'}$ and public attribute $u$.
Figure~\ref{fig:transformation} illustrates the transfer vector in a three-dimensional latent space. 
The markers show only the latent representations of embeddings with public attribute $u$.
Circles and squares are representations of data embeddings 
with private attribute class $i^{'}$ and class $i$, respectively.
The mean latent representation is depicted by a cross in each case.
Once the transfer vector is found, it can be applied to modify the private attribute 
of a given data embedding as described in Algorithm~\ref{algorithm:1}. \textcolor{red}{In Step~6, 
we finally use the corresponding decoder network 
to construct the obfuscated version of the input data embedding, $\hat{x}_k$.}

\textcolor{red}{We note that the \texttt{Modify} function can be either deterministic or probabilistic.}
When the private attribute is binary, 
the deterministic modification converts one class label to the other one at all times.
When the private attribute class is not binary, an arbitrary bijective function can be used. 
\textcolor{red}{In the case of probabilistic transformation, 
the mean manipulation is performed with probability of 0.5.}
Specifically, for each data embedding we decide whether to perform the mean manipulation
based on a cryptographically secure stream of pseudo random numbers.
We use the \emph{CPRNG Secrets}\footnote{\href{https://docs.python.org/3/library/secrets.html}{https://docs.python.org/3/library/secrets.html}.} python module to generate the random numbers. 
\section{Evaluation Results}
\label{eval}
\subsection{Datasets}
To evaluate the efficacy of the proposed anonymization technique, 
we use two publicly available HAR datasets, namely MotionSense and MobiAct.

\subsubsection{MotionSense}
This dataset contains measurements of accelerometer and gyroscope sensors~\cite{malekzadeh19}. 
This dataset is collected from iPhone~6s using the Sensing Kit framework~\cite{katevas2016sensingkit}. 
It contains data from 24 subjects (14 males and 10 females), 
each performing 15 trials that include 6 different activities. 
The activities include climbing stairs up and down, walking, jogging, sitting, and standing. 
The subjects' age, height, and weight cover a wide range of values. 
The dataset is collected at 50~Hz sampling rate and 
each sample contains 12 features including attitude (roll, pitch, yaw), 
gravity, rotation rate, and user acceleration in three dimensions, $x$, $y$, and~$z$.


We use windows of 128~samples, each corresponding to time series data sampled over $2.56$ seconds. 
These windows are moved with strides of 10~samples. 
This is the same configuration used in~\cite{malekzadeh19}. 
We use this configuration so that we can have a fair comparison between the anonymization results. 

\textcolor{red}{
From the 6 activities mentioned earlier, standing and sitting have quite similar features. 
In fact the only distinction between standing and sitting activities 
in terms of IMU readings is the position of the smartphone in the users’ pocket, 
which is vertical in one case and horizontal in the other. 
Hence, in~\cite{malekzadeh19}, these activity labels are merged into one label, namely standing. 
But even after combining these activities, we do not have enough data for the new activity
to perform sensitive inferences, e.g., gender identification. 
Thus, following~\cite{malekzadeh19} where the MotionSense dataset was originally introduced,
we ignore standing and sitting activities.}
We use trials $11$, $12$, $13$, $14$, $15$, and $16$ from this dataset to form our test set. 
We treat each of these 4 activity classes as a public attribute class, $u$, 
and each of the 2 gender classes, labelled $0$ and $1$, as a private attribute class, $i$.

\subsubsection{MobiAct}
This dataset is comprised of smart phone sensor readings.
It includes different falls in addition to various activities~\cite{vavoulas2016mobiact}. 
This dataset is larger than MotionSense in terms of the number of activities and participants. 
There are $66$ participants in MobiAct performing $12$ daily lives activities, 
such as running, jogging, and going up or down the stairs. 
Besides these $12$ activities, the subjects perform $4$ different types of falls. 
From the $66$ participants, we use only $37$ participants 
to create a relatively balanced dataset in terms of the number of female and male participants. 
In particular, there are $17$ females (gender $1$) and $20$ males (gender $0$) in the subsampled dataset. 

To investigate if our anonymization technique can deal with a non-binary private attribute, 
we bin the recorded weights into three classes.
Specifically, we label subjects that weigh less than or equal to $70$~kg as $0$, 
those who weigh between $70$ and $90$~kg as $1$,
and those who weigh more than $90$~kg as $2$.

From the large list of activities in this dataset, we select four activities 
for which all participants have representative data and sensor readings. 
These activities are walking, standing, jogging, and climbing stairs up,
which are referred to as WAL, STD, JOG, and STU, respectively.
Since not all activities are performed in more than one trial, 
we use a subject-based train-test split. 
We use $80\%$ of the available data for training and the remaining $20\%$ for test.

\subsection{Hyperparameters Tuning}
We perform grid search to tune hyperparameters $\alpha$ and $\beta$.
We consider a range of values for $\alpha$ and $\beta$ for each dataset
and choose the values that result in better anonymization performance.
Concretely, these parameters should result in a lower average loss across all attribute-specific VAEs. \textcolor{red}{Note that hyperparameters are tuned prior to the deployment of the models. 
In other words, tuning hyperparameters is a one-time cost and should not affect the running time of the anonymization pipeline.}

For MotionSense, we assign $16$ pairs of values to $\alpha$ and $\beta$, 
setting $\alpha$ to $0.5$, $1$, $2$, and $3$, and $\beta$ to $1$, $2$, $3$, and $4$.
We find that the best performance is achieved when $\alpha=2, \beta=2$.
For MobiAct, we select the most suitable weights for $\alpha$ and $\beta$ in the same fashion. 
In this case, our empirical results suggest that we can further increase 
the weight of KL-divergence, $\beta$, to $6$. 
We consider $0.5$, $1$, $2$, and $3$ for $\alpha$, and $2$, $4$, $5$, and $6$ for $\beta$. 
The best anonymization performance is attained when $\alpha=2, \beta=5$ for gender anonymization 
and $\alpha=1, \beta=5$ for weight group anonymization.

\subsection{Anonymization Results: MotionSense}
We first use our proposed anonymization technique to obfuscate sensor data in the MotionSense dataset.
Here the gender identity of the user (male or female) 
is the private attribute and their activity is the public attribute.
Thus, gender identification is the sensitive inference and 
activity recognition is the desired inference.
We report the anonymization results when we modify the latent representation of input data
in a deterministic and probabilistic fashion.

\textcolor{red}{It is worth mentioning that since the private attribute is binary, 
if the sensitive inference accuracy can be reduced to around $0\%$ 
by modifying latent representations in a deterministic fashion,
we can achieve almost the same accuracy as random guess ($50\%$) 
by modifying the representations with $0.5$ probability.
In general, when we use deterministic modifications,
the lower the accuracy of a sensitive inference is, 
the closer we can get to $50\%$ accuracy by introducing randomness 
through probabilistic modifications.
Note that this would be the ideal outcome because it confuses the training algorithm
used by the attacker in the private attribute re-identification attack to the greatest extent, 
thereby offering protection against it.}


\subsubsection{Anonymization with Deterministic Modification}
To evaluate our technique with a deterministic modification, 
we use the same architecture for the human activity recognition and gender identification models 
as our previous work~\cite{hajihass2020latent}. 
These models are Multilayer Perceptron (MLP) neural networks 
which are discussed in Section~\ref{implementation}.

\begin{figure}
     \centering
     \begin{subfigure}[t]{0.5\textwidth}
         \centering
         \includegraphics[width=\textwidth]{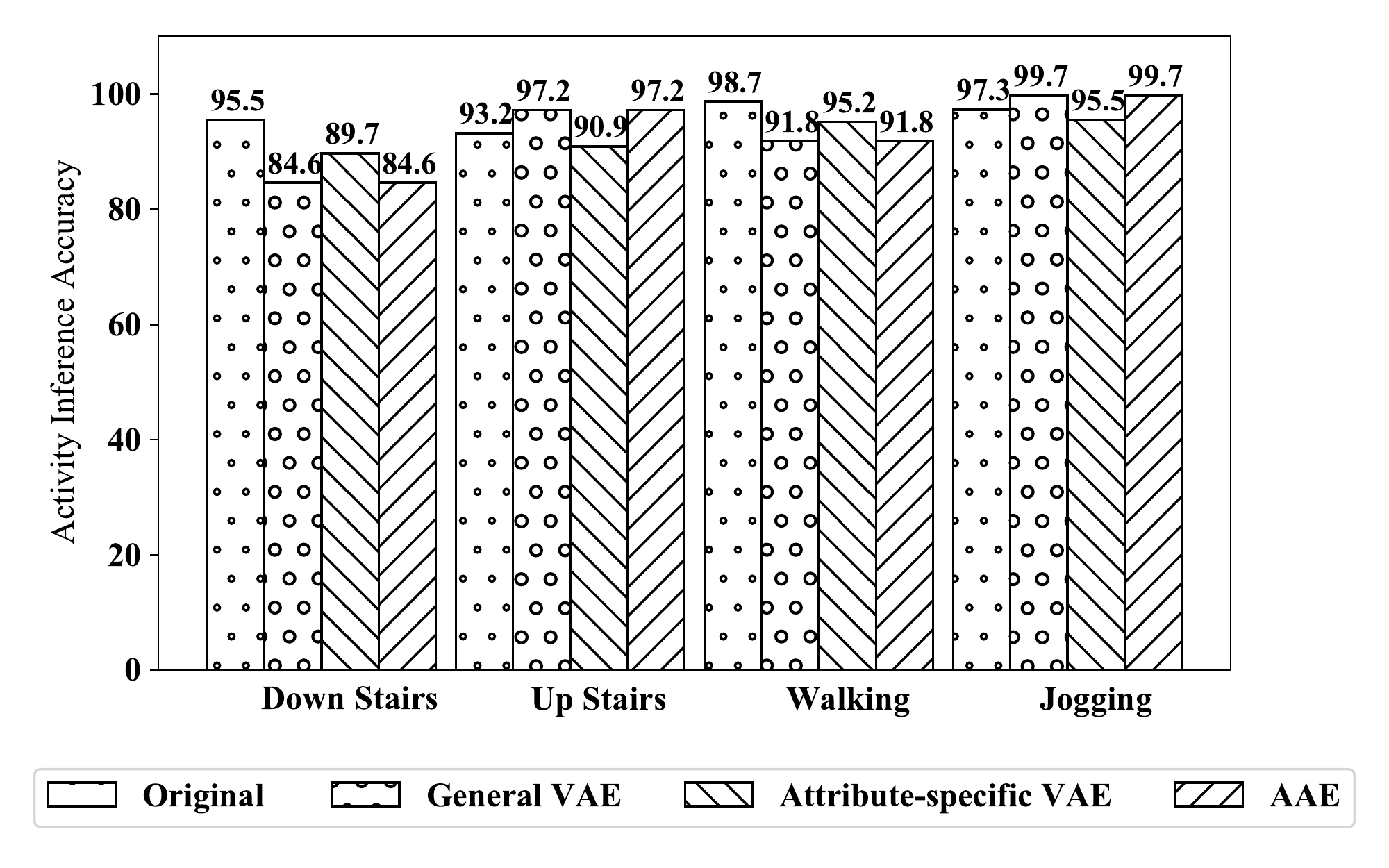}
         \caption{Activity inference}
         \label{fig:detact}
     \end{subfigure}
     ~
     ~
     \begin{subfigure}[t]{0.5\textwidth}
         \centering
         \includegraphics[width=\textwidth]{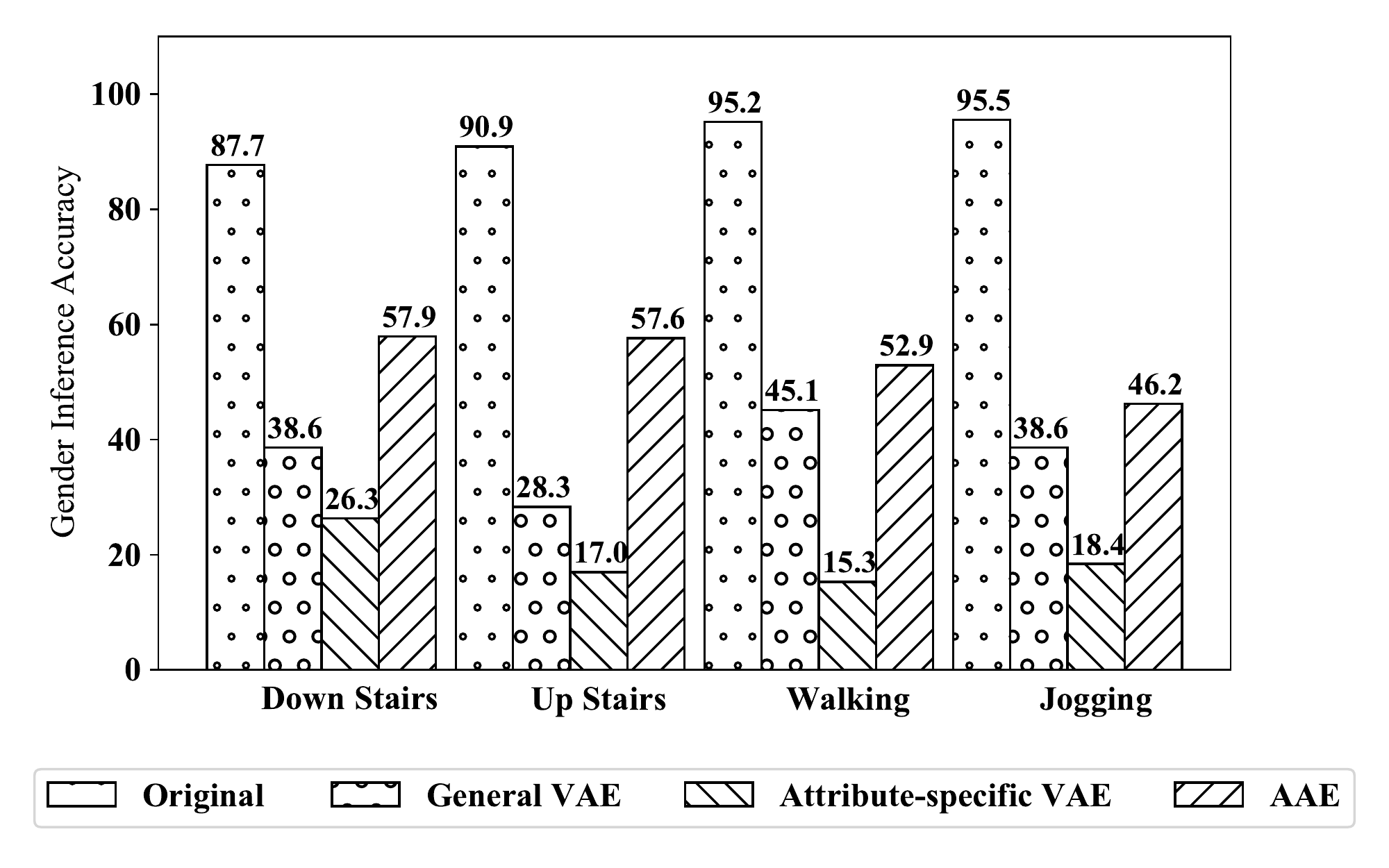}
         \caption{Gender inference}
         \label{fig:detgend}
     \end{subfigure}
        \caption{Accuracy of inference models on the anonymized MotionSense test set 
        for two-class gender anonymization. \textcolor{red}{Here activity is regarded as the public attribute 
        that is being inferred, while gender is the private attribute that must be concealed.}}
        \label{fig:three graphs1}
\end{figure}

We evaluate our attribute-specific VAE models on the test set. 
We obtain the mean values of latent representations 
for each private attribute class (gender identity) and each public attribute class (activity type). 
These mean values are denoted ${\bar{z}}^{i}_{u}$ in Algorithm~\ref{algorithm:1}
and are calculated using the data in the training set.
We use these mean values to perform mean manipulation 
in the latent space of a VAE as described in Algorithm~\ref{algorithm:1}.
Figures~\ref{fig:detact} and~\ref{fig:detgend} show the accuracy of desired and sensitive inferences on 
the data anonymized by the proposed technique using deterministic modifications. 
\textcolor{red}{
We first compare the accuracy of using a general VAE to learn and modify latent representations, 
i.e., the method introduced in~\cite{hajihass2020latent}, 
with the accuracy of using multiple attribute-specific VAEs to learn and modify latent representations,
i.e., the method proposed in this work.
We see that attribute-specific VAEs can further reduce the privacy loss,
while maintaining the data utility to a comparable extent. 
This supports our claim that a general VAE lacks enough learning capacity 
to represent embeddings from all public attributes and leads 
to degradation in anonymization performance when we use deterministic modifications.}
As it can be seen from Figure~\ref{fig:detgend}, 
our anonymization technique also outperforms Anonymizing Autoencoder (AAE)~\cite{malekzadeh19} 
in terms of privacy loss.
Note that the proposed anonymization technique does not know the true private and public attributes of the user,
and must predict them using pretrained models as depicted in Figure~\ref{fig:modvae}.

\begin{table}[htbp]
  \centering
  \caption{Accuracy of activity and gender recognition models on the anonymized MotionSense test dataset.
  The results are reported separately for each activity label (first column). 
  The number of data embeddings is specified for each activity (last column).}
    \begin{tabular}{c|c|c|c|c|c}
    Activity Class & Act. Bef & Act. After & Gen. Before & Gen. After & nb. Embeddings\\ \toprule
    Down stairs     & 95.58\%  & 89.73\% & 87.67\% & 26.25\% & 1.9k\\
    Up stairs     & 93.19\% & 93.47\% & 90.86\% & 17.03\% & 2.5k\\
    Walking     & 98.71\% & 98.49\% & 95.16\% & 15.34\% & 6.2k\\
    Jogging     & 97.28\% & 97.28\% & 95.50\% & 18.39\% & 2.7k\\ \bottomrule
    \end{tabular}%
  \label{tab:addlabel}%
 \centering
\end{table}%

Table~\ref{tab:addlabel} shows the accuracy of the activity and gender identification models
before and after anonymizing data in the test set for each activity class. 
These models are also used as the inference models in Algorithm~\ref{algorithm:1} 
for predicting public and private attribute classes.
We see that anonymizing with deterministic modifications makes possible 
up to $78.21\%$ reduction in the gender identification accuracy (from $93.35\%$ to $17.84\%$). 
Notice that these results are the weighted average of inference accuracy levels
based on the number of embeddings available for each activity, as shown in Table~\ref{tab:addlabel}.
Moreover, the public attribute inference model is improved by roughly $1\%$ across different activities. 
In our previous work~\cite{hajihass2020latent} we achieved the gender identification accuracy of $39.62\%$ 
(on average) after anonymization.
We reduced this accuracy to $17.84\%$ in this work,
while increasing the activity recognition accuracy noticeably; 
we attribute this positive result to the use of attribute-specific VAEs 
which allows for learning better representations in a highly imbalanced dataset. 



Suppose the attacker has access to $20\%$ of the sensor data obfuscated 
using our technique along with the corresponding private attribute.
The $20\%$ data is sampled from the training set uniformly at random, 
and is used to train a model to perform the re-identification attack.
We observe that the attacker can achieve more than $95\%$ accuracy 
in the gender identification task despite using our anonymization technique.
We attribute this to the deterministic nature of modifications in the latent space
that enables the attacker to trivially learn a model to re-identify the gender.

\subsubsection{Anonymization with Probabilistic Modification}
\begin{figure}[t]
\centering
  \includegraphics[scale=.5]{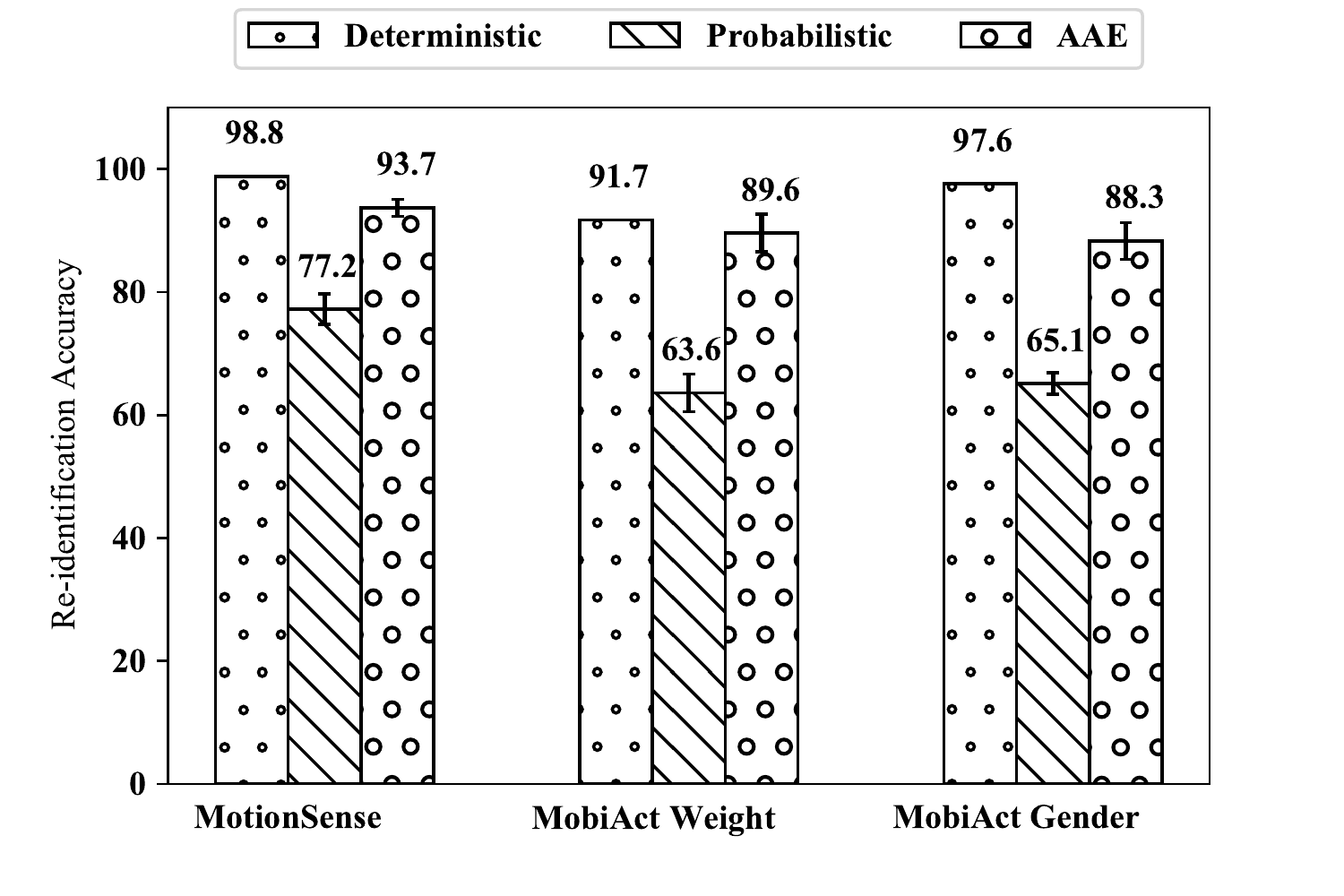}
  \caption{\textcolor{red}{The private attribute re-identification accuracy 
  (indicating the success of the re-identification attack)
  when sensor data is obfuscated using Anonymizing Autoencoders~\cite{malekzadeh19} 
  and our proposed technique with deterministic and probabilistic modifications.}}
  \label{fig:prob_reid}
\end{figure}

\textcolor{red}{We now show that by modifying the latent representations in a probabilistic fashion, 
the re-identification attack can be prevented to a great extent.
Similar to the previous section, we assume the attacker has access to $20\%$ of the obfuscated data 
along with the true private attribute associated with this data, 
and uses this data to train a model that re-identifies the gender of the user.
But this time we anonymize the data using the proposed anonymization technique
with probabilistic modifications. 
We evaluate the success of the re-identification attack 
by looking at the accuracy of the model trained by the attacker.
To account for the randomness that may arise from training, anonymization, and sampling process, 
we consider $20$ independent runs 
and plot the average and standard deviation of the accuracy results in Figure~\ref{fig:prob_reid}.
It can be readily seen that our approach with a deterministic modification
and AAE~\cite{malekzadeh19} (our baseline) are susceptible to the re-identification attack 
because the attacker can re-identify the private attribute (gender) 
with $98.8\%$ and $93.7\%$ accuracy, respectively.
However, the accuracy of the private attribute re-identification model reduces to $77.2\%$
using a simple probabilistic modification. 
This implies that the re-identification attack is less successful in this case.}

\textcolor{red}{We argue that the major difference between our anonymization technique and 
adversarial model-based techniques, for example AAE, is that 
we attempt to reduce the accuracy of sensitive inferences to the level of random guessing
by introducing randomness in the process of modifying latent representations. 
But adversarial model-based techniques achieve this through the use of a specific model for adversarial training.
As a result, these techniques cannot prevent the re-identification attack
because the attacker model can be different from the model that was originally used for adversarial training.
In this work, the model trained by the attacker is either a convolutional or MLP neural network. 
We assume it has the same architecture as the gender identification model.}

\subsection{Anonymization Results: MobiAct}
We use the MobiAct dataset for two-class gender and multi-class weight group anonymization. We evaluate our anonymization technique on this dataset besides MotionSense because it allows us to study the case where the private attribute is non-binary.
The public attribute inference model estimates the daily activity of each user, 
while the private attribute inference model aims to infer the gender identity (2 classes) of the user in one case
and the weight group (3 distinct classes) of the user in the other case.
We use Convlutional Neural Network (CNN) models as our activity, gender, and weight group inference models.
More details about these inference models are given in Section~\ref{implementation}.

\subsubsection{Anonymization with Deterministic Modification}
\begin{figure}
     \centering
     \begin{subfigure}[t]{0.5\textwidth}
         \centering
         \includegraphics[width=\textwidth]{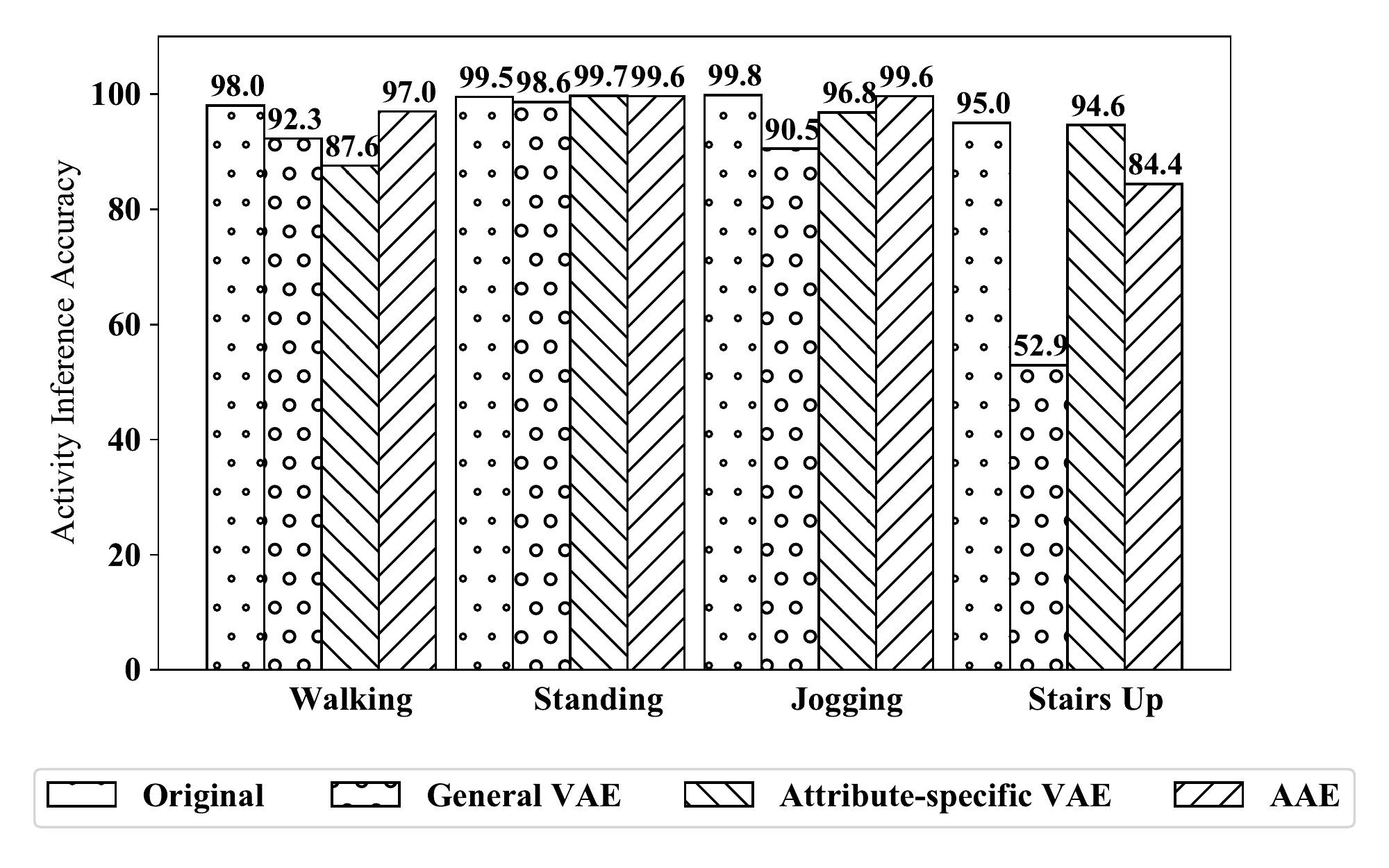}
         \caption{Activity inference}
         \label{fig:act}
     \end{subfigure}
     ~
     ~
     \begin{subfigure}[t]{0.5\textwidth}
         \centering
         \includegraphics[width=\textwidth]{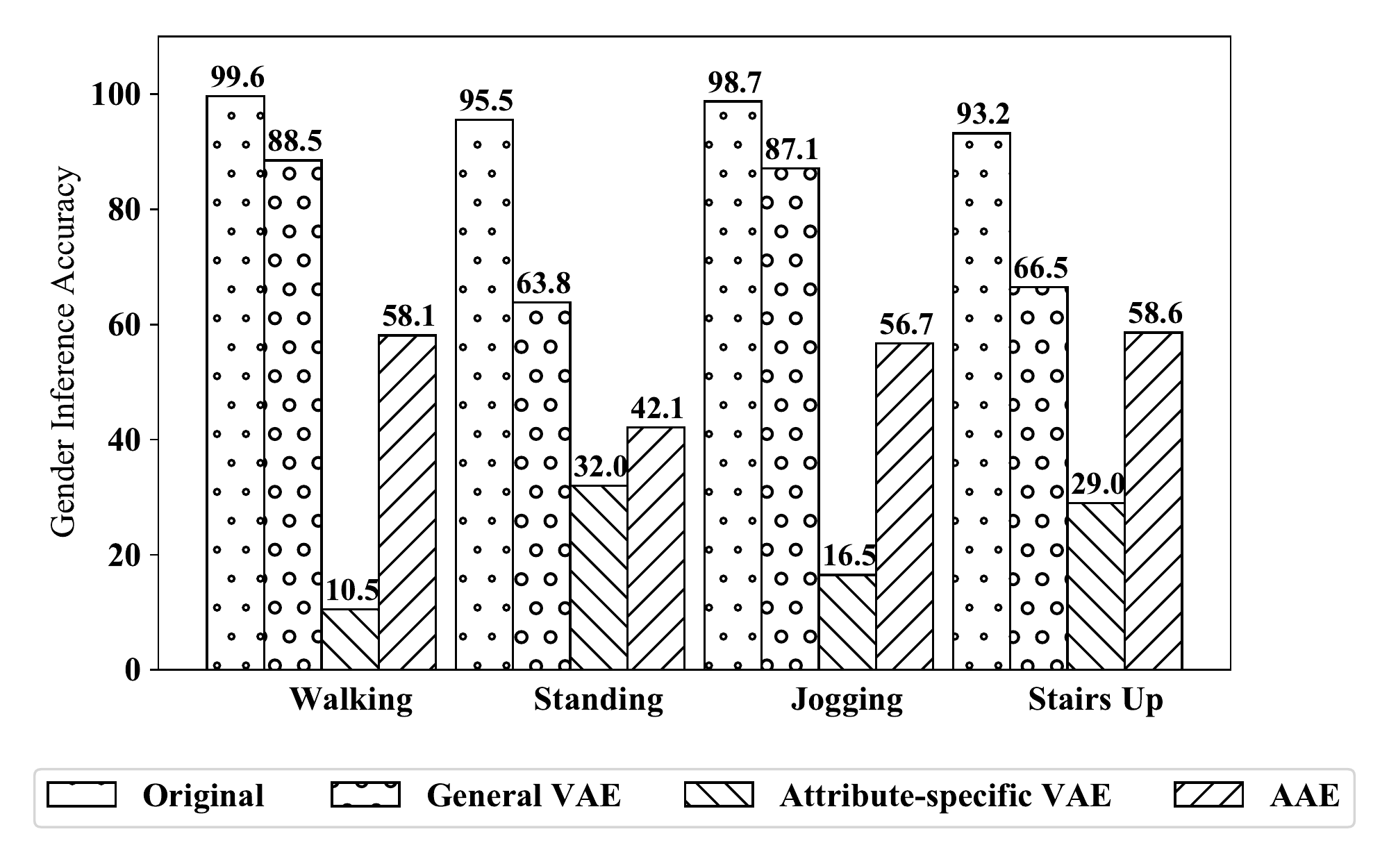}
         \caption{Gender inference}
         \label{fig:act_gen}
     \end{subfigure}
        \caption{Accuracy of inference models on the anonymized MobiAct test set for two-class gender anonymization. \textcolor{red}{Here activity is regarded as the public attribute 
        that is being inferred, while gender is the private attribute that must be concealed.}}
        \label{fig:three graphs2}
\end{figure}

\begin{figure}
     \centering
     \begin{subfigure}[t]{0.5\textwidth}
         \centering
         \includegraphics[width=\textwidth]{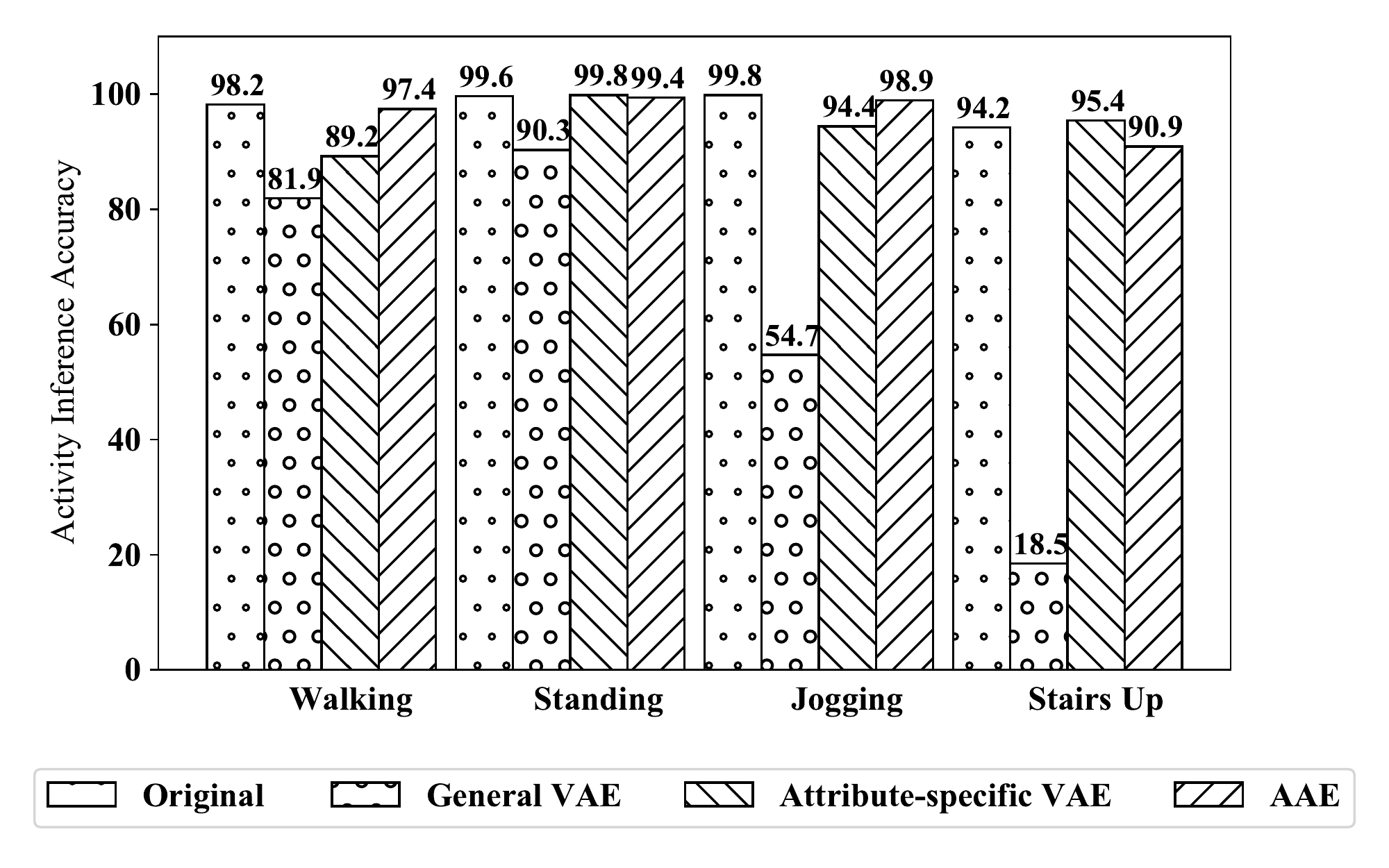}
         \caption{Activity inference}
         \label{fig:act_weight}
     \end{subfigure}
     ~
     ~
     \begin{subfigure}[t]{0.5\textwidth}
         \centering
         \includegraphics[width=\textwidth]{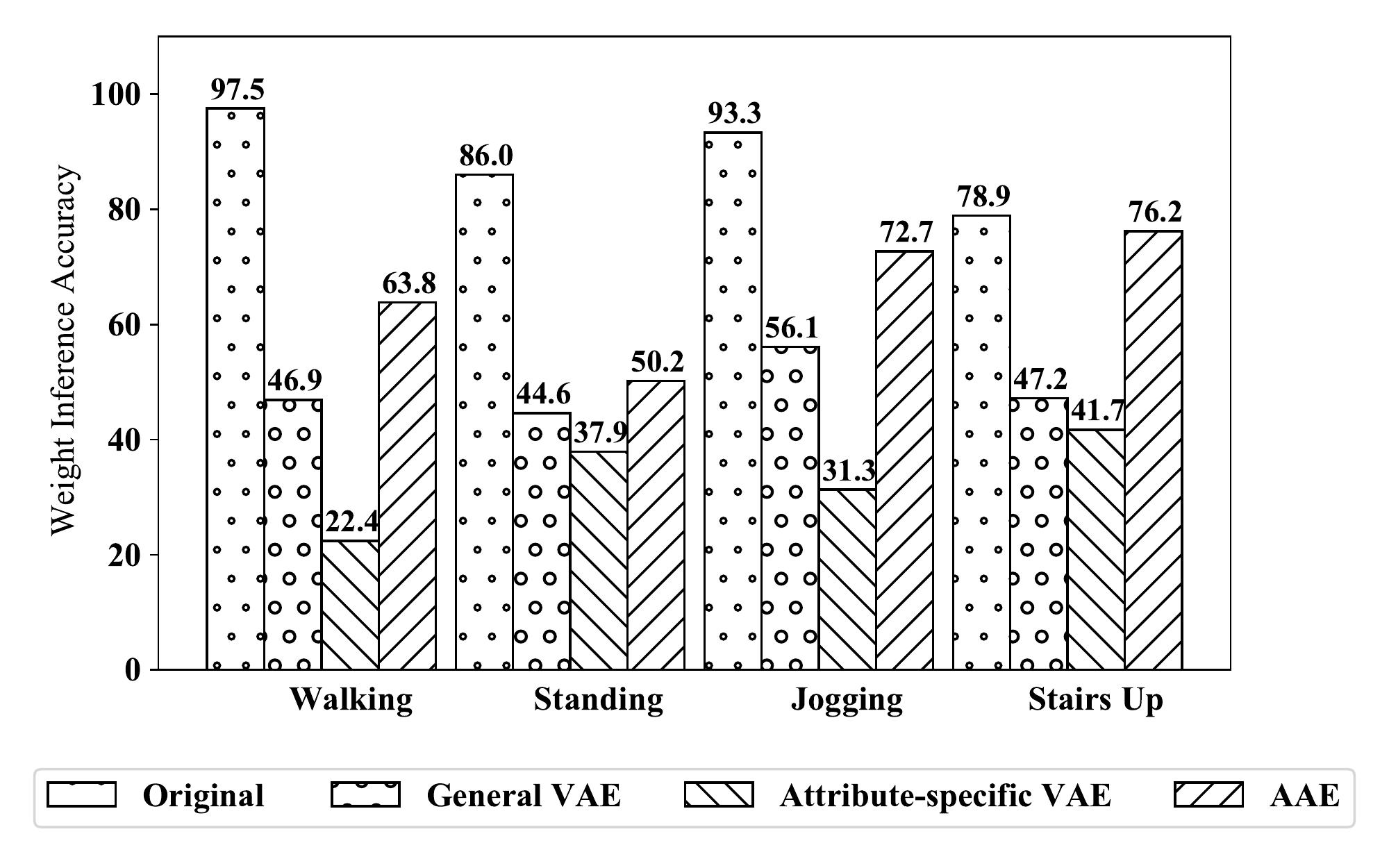}
         \caption{Weight group inference}
         \label{fig:weight}
     \end{subfigure}
        \caption{Accuracy of inference models on the anonymized MobiAct test set for the ternary class weight group anonymization. \textcolor{red}{Here activity is regarded as the public attribute 
        that is being inferred, while weight group is the private attribute that must be concealed.}}
        \label{fig:three graphs3}
\end{figure}

The anonymization is first performed by modifying the private attribute in a deterministic fashion. 
We first discuss the two-class gender anonymization results. 
Figures~\ref{fig:act} and~\ref{fig:act_gen} show respectively 
the accuracy of the activity and gender inference models on the test set. 
Compared to our anonymization technique, AAE~\cite{malekzadeh19} performs worse, 
yielding a higher gender identification accuracy. 
Note that the activity indices are different from MotionSense. 
For weight group anonymization, we modify weight group attribute as follows: 
from $0$ to $1$, from $1$ to $2$, and from $2$ to $0$. 
This is an arbitrary mapping and can be changed, but the point is that it is done in a deterministic fashion. 
The accuracy of activity and weight group inference models on the test set 
is depicted in Figures~\ref{fig:act_weight} and~\ref{fig:weight}, respectively\footnote{Since 
we did not have the original implementation of AAE~\cite{malekzadeh19} for the MobiAct dataset, 
we built our own AAE model and tested it on this dataset.}. 
\textcolor{red}{By inspecting these figures, 
we can contrast the performance of the general VAE with that of attribute-specific VAEs. 
As it can be seen, using the attribute-specific VAEs trained specifically for each public attribute 
yields better results than using the general VAE in terms of utility and privacy loss.}

Tables~\ref{tab:table2} and~\ref{tab:table3} show results of the deterministic anonymization 
for gender and weight group private attribute classes in the MobiAct dataset. 
In the case of the two-class gender anonymization,
we conclude 
that the activity detection accuracy is dropped by around $5\%$ 
while the gender inference accuracy is decreased by $76.36\%$ (from $95.52$ to $21.16\%$). 
Turning our attention to the three-class weight anonymization, 
our results indicate that the activity detection accuracy is dropped only slightly ($4.38\%$). 
This is while the weight group inference accuracy is decreased by $61.27\%$ (from $91.64$ to $30.37\%$).



\begin{table}[htbp]
  \centering
  \caption{Accuracy of the activity and gender Inference in MobiAct test dataset gender anonymization.
  The results are reported separately for each activity label (first column) with the number of data embeddings in each case.}
    \begin{tabular}{c|c|c|c|c|c}
    Activity Class & Act. Before & Act. After & Gen. Before & Gen. After & nb. Embeddings \\ \toprule
    Walking & 98.00\% & 87.57\% & 99.58\% & 10.53\% & 42.9k \\
     Standing & 99.52\% & 99.67\% & 95.45\% & 31.98\% & 43.2k \\
    Jogging & 99.78\% & 96.75\% & 98.72\% & 16.51\% & 4.2k \\
    Stairs up & 95.03\% & 94.63\% & 93.22\% & 28.99\% & 1k \\ \bottomrule
    \end{tabular}%
  \label{tab:table2}%
\end{table}%

\begin{table}[htbp]
  \centering
  \caption{Accuracy of the activity and weight Inference in MobiAct test dataset weight anonymization.
  The results are reported separately for each activity label (first column) with the number of data embeddings in each case.}
    \begin{tabular}{{c|c|c|c|c|c}}
    Activity Class & Act. Before & Act. After & Weight Before & Weight After & nb. Embeddings\\ \toprule
    Walking & 98.16\% & 89.15\% & 97.48\% & 22.41\% & 42.9k \\
    Standing & 99.58\% & 99.75\% & 85.98\% & 37.93\% & 43.2k \\
    Jogging & 99.80\% & 94.43\% & 93.30\% & 31.25\% & 4.2k \\
    Stairs up & 94.22\% & 95.36\% & 78.88\% & 41.69\% & 1k \\ \bottomrule
    \end{tabular}%
  \label{tab:table3}%
\end{table}%

\subsubsection{Anonymization with Probabilistic Modification}
We now investigate the efficacy of our anonymization technique with probabilistic modifications.
We use CNNs as inference models for the weight group and gender;
these models are described in Section~\ref{implementation}.

We first focus on the inference accuracy of a gender re-identification model.
As before, we assume the attacker has access to $20\%$ of the obfuscated data 
along with the true private attribute associated with this data, 
and uses this data to train a model to re-identify the gender of the user.
The average and standard deviation of the accuracy of this attacker model across $20$ independent runs,
when sensor data is obfuscated using our anonymization technique with probabilistic modifications
are depicted in Figure~\ref{fig:prob_reid} (the 3 bars in the right side).
We can see that the re-identification attack is less successful in this case than 
when we use deterministic modifications or AAE to obscure the private attribute.
Next, we turn our attention to the success of the re-identification attack 
when the weight group is the private attribute. 
We conduct the same study using our anonymization technique with probabilistic modifications.
The results obtained in $20$ independent runs are shown in Figure~\ref{fig:prob_reid} (the 3 bars in the middle).
Similar to the gender anonymization case, 
using the proposed anonymization technique with probabilistic modifications 
prevents the attacker from re-identifying the weight group to a great extent 
compared to the other two techniques.



\section{Implementation Details and Practical Considerations}
\label{implementation}

Performing anonymization at the edge is essential for real-world applications 
since the user may not trust cloud servers to operate on their sensor data before it is obfuscated.
However, running the proposed anonymization technique is a compute intensive task.
In this section, we measure the running time of the adversarial model-free anonymization technique 
on a Raspberry~Pi~3~Model~B to understand if it can run in real time on a resource-constrained edge device.
\textcolor{red}{The Raspberry~Pi~3~Model~B has a 1.2GHz quad core CPU with 1GB of memory;
the implementation of our anonymization technique uses its CPU only.
Raspberry~Pi is used in this work to represent a low-power edge device. 
Since it runs Raspbian, which is a Linux-based operating system, 
it is straightforward to port the current implementation of PyTorch.
This enables us to train various neural network models.}

There are two factors that affect the time budget we have 
for real-time execution of the proposed anonymization technique.
The first factor is the rate at which new sensor data becomes available;
this is usually the same as the lowest sampling rate of the respective sensors.
The second one is the embedding size that we use for anonymization.
If the total running time of the proposed anonymization technique exceeds this time budget, 
data anonymization cannot be carried out in real time.
The total running time of the proposed anonymization technique is the sum of 
the running times of the pretrained classifiers for private and public attributes, 
probabilistic encoder, linear transformation, and probabilistic decoder.

The sampling rate of sensors is $20$Hz in the MobiAct dataset. 
Hence, the Inertial Measurement Unit (IMU) readings become available every twentieth of a second. 
We generate embeddings with windows of $128$ samples and strides of $10$ samples,
which means that after the first embedding is generated, 
a new embedding is generated every half a second ($500$ milliseconds).
Similarly, the sampling rate of sensors is $50$Hz in the MotionSense dataset.
In this case we generate data embeddings with windows of $128$ samples and strides of $10$ samples.
Hence, a new embedding becomes available every $200$ milliseconds.
With this calculation, the total time budget we have for the
real-time execution of our anonymization technique can be found.

We remark that there is a small difference in the way that embeddings are generated for each dataset.
For the inference models to achieve high accuracy in the MobiAct dataset, 
the value of each coordinate should be utilized instead of the magnitude of the sensor readings. 
Thus, when considering the data produced by gyroscope and accelerometer sensors, 
$6$ $\times$ $128$ coordinate readings must be stored in each embedding 
rather than $2$ $\times$ $128$ magnitude readings. 
As a result, the size of the inference models and the size of the VAEs in MobiAct 
are bigger than those in MotionSense.
Keep in mind that using smaller-size public attribute-specific VAEs 
is beneficial for the real-time execution of our anonymization technique. 
A general VAE has 12~times more trainable weights than the 
total number of trainable weights of all attribute-specific VAEs.

\begin{figure}
     \centering
     \begin{subfigure}[t]{0.46\textwidth}
         \centering
         \includegraphics[width=\textwidth]{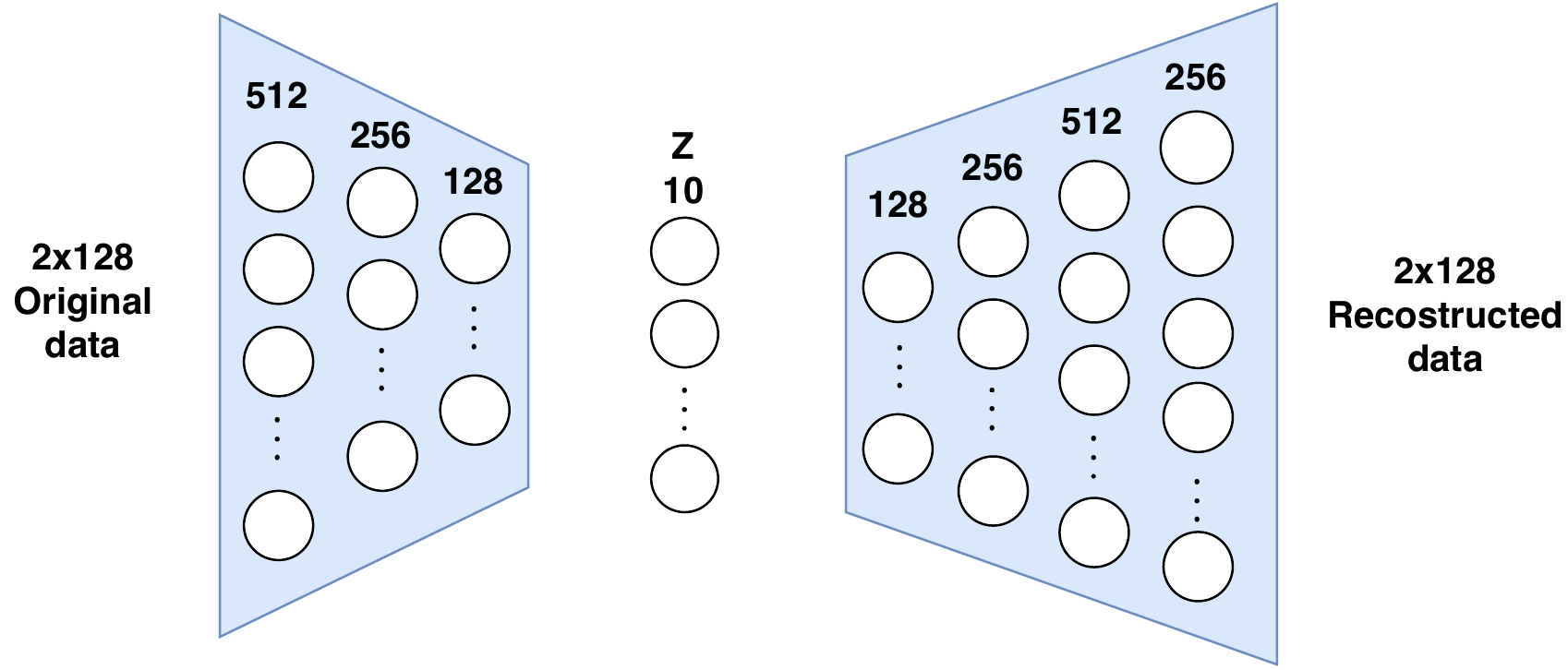}
         \caption{An attribute-specific VAE for MotionSense}
         \label{fig:vae_motion}
     \end{subfigure}
     ~
     ~
     \begin{subfigure}[t]{0.45\textwidth}
         \centering
         \includegraphics[width=.8\textwidth]{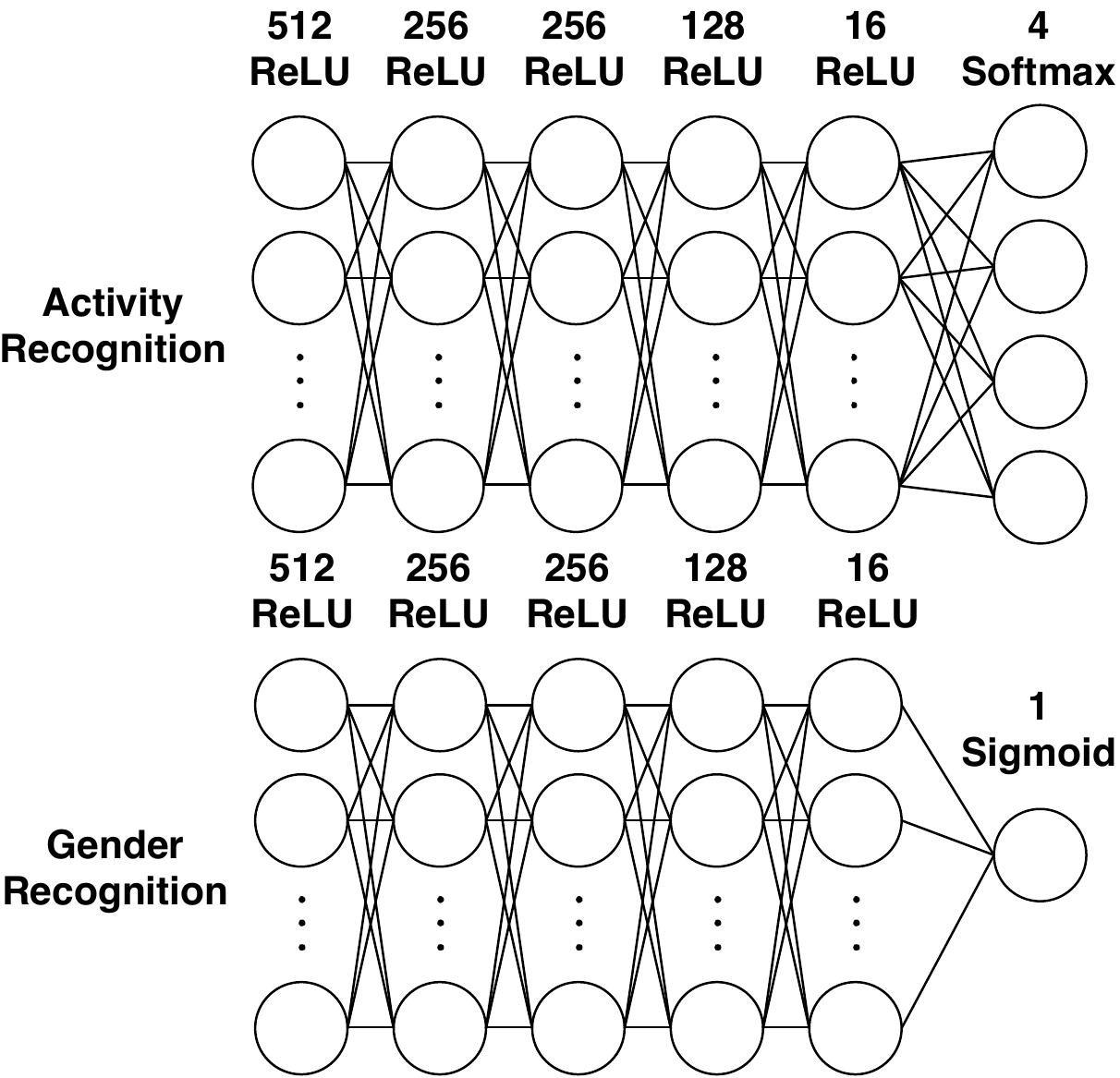}
         \caption{MLPs for activity and gender identification}
         \label{fig:motion_mlp}
     \end{subfigure}
        \caption{Neural network architecture and the number of neurons in each layer for the MotionSense dataset.}
        \label{fig:motion}
\end{figure}

\textcolor{red}{We measure the running time of each step of Algorithm~\ref{algorithm:1} on the Raspberry~Pi on three separate runs and report the average value of these independent runs.
The first step (Step~1) is to use the two classifiers to identify public and private attribute labels.
Since we have different classifiers for activity, gender, and weight group,
we report their running times separately in the following sections.
Furthermore, we report the running time of each attribute-specific encoder and decoder network,
which are called in Step~2 and Step~6 of the algorithm.
Our evaluation shows that the total running time of Steps~3-5 of the algorithm,
where we transform a latent representation, is several orders of magnitude smaller than the other steps.
Thus, we do not consider this in our calculations.
We also assume that the classifiers and VAEs are loaded in memory only once.}

\subsection{MotionSense dataset}
In the MotionSense dataset, it takes $0.72$ and $0.71$ milliseconds
to run the classifiers to identify the public and private attributes 
of one input data embedding, respectively. \textcolor{red}{These results are the reported average of 3 independent and separate runs on the Raspberry Pi 3 device.}
In addition to that, it takes up to $2.4$ and $1.8$ milliseconds 
for the probabilistic encoder and decoder of the VAEs to run, respectively. 
Thus, in the worst case, it takes about $5.63$ milliseconds in total to anonymize one embedding. 
This result is shown in Table~\ref{fig:MotionSense_time} and the maximum running time is printed in bold.
Considering the 200-millisecond time budget we have, 
we conclude that it is feasible to perform real-time data anonymization on a typical edge device. 

Note that for the sake of comparison with~\cite{hajihass2020latent},
MLP models are used as classifiers (for activity and gender identification). 
Figure~\ref{fig:motion} shows the architecture of these MLP models and the VAE.

\begin{table}[h]
\caption{The running time of different components of the proposed anonymization technique in the gender anonymization task in MotionSense.}
\begin{tabular}{c|c|c|c|c|c}
Model                & Batch Sizes & Type & nb. Embeddings & Time (s) & \begin{tabular}[c]{@{}c@{}}Time/\\ Embedding (s)\end{tabular}       \\ \toprule
Activity Recognition & 256         & MLP  & 21,210        & 15.3633 & \textbf{0.00072} \\
Gender Recognition   & 256         & MLP  & 21,210
        & 15.0095 & \textbf{0.00071} \\
Prob. Encoder 0      & 256         & MLP  & 1,947
         & 4.67   & \textbf{0.00239} \\
Prob. Decoder 0      & 256         & MLP  & 1,947
         & 3.4948 & \textbf{0.00179} \\
Prob. Encoder 1      & 256         & MLP  & 2,495
         & 5.5973 & 0.00224                \\
Prob. Decoder 1      & 256         & MLP  & 2,495
         & 4.2394  & 0.00169                \\
Prob. Encoder 2      & 256         & MLP  & 6,225
           & 10.7227 & 0.00172                \\
Prob. Decoder 2      & 256         & MLP  & 6,225
           & 8.8715  & 0.00143                \\
Prob. Encoder 3      & 256         & MLP  & 2,686
            & 5.8151  & 0.00216 \\
Prob. Decoder 3      & 256         & MLP  & 2,686
            & 4.4657 & 0.00166                \\ \bottomrule
\end{tabular}
\label{fig:MotionSense_time}
\end{table}

\subsection{MobiAct dataset}
We now present the running times of weight-group and gender anonymization tasks in the MobiAct dataset. 
Let us focus on the binary gender anonymization task first. \textcolor{red}{On an average of three separate runs on the Raspberry Pi 3}, we find that it takes $49.4$ and $49.3$ milliseconds to run the classifiers 
to identify the public and private attributes of one input data embedding, respectively.
Also, in the worst case, the probabilistic encoder of the VAE takes about $6.53$ milliseconds 
and its probabilistic decoder takes about $6.32$ milliseconds to process one data embedding. 
These numbers add up to $111.55$ milliseconds, 
which is the total time it takes to anonymize one data embedding in the worst case. 

\begin{figure}
     \centering
     \begin{subfigure}[t]{0.43\textwidth}
         \centering
         \includegraphics[width=\textwidth]{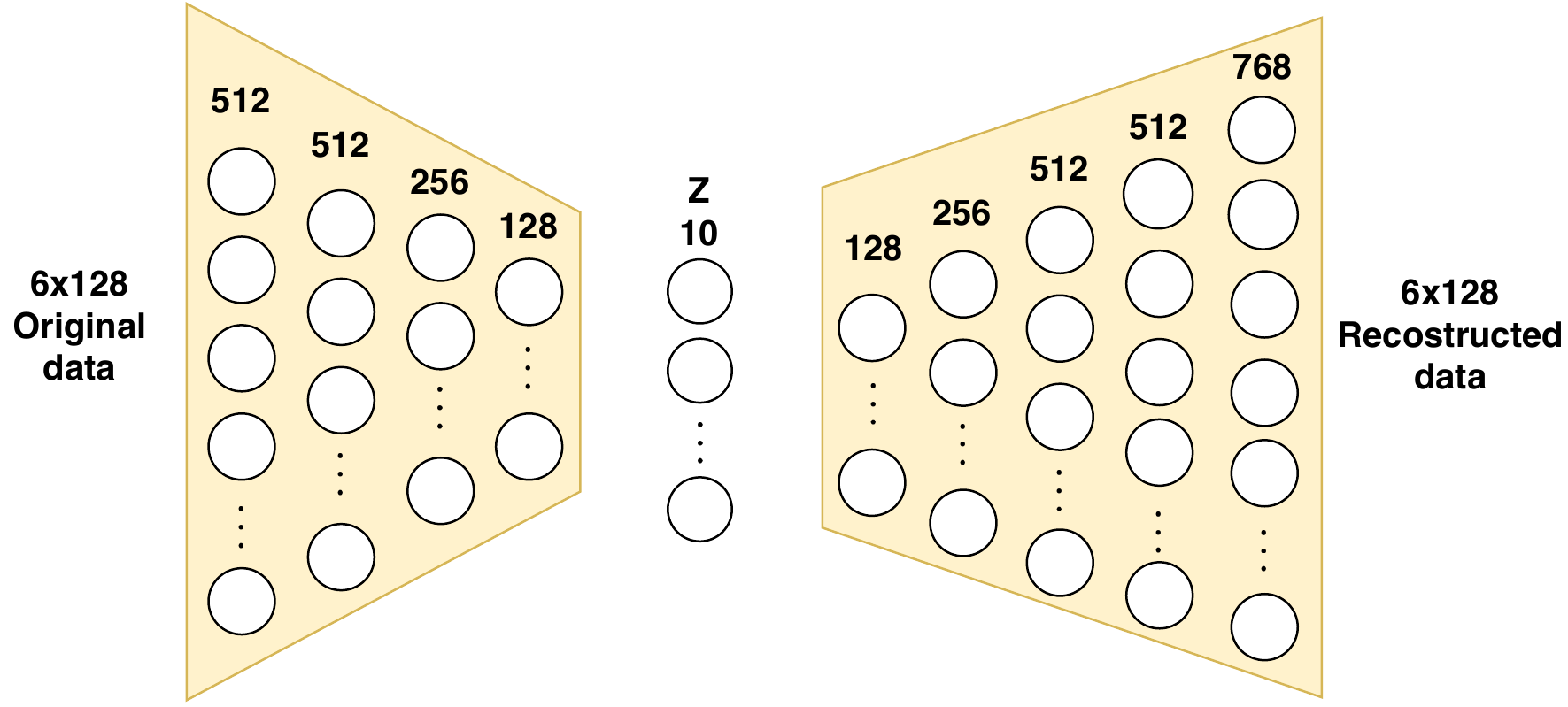}
         \caption{An attribute-specific VAE in MobiAct}
         \label{fig:vae_mobi}
     \end{subfigure}
     ~
     ~
     \begin{subfigure}[t]{0.53\textwidth}
         \centering
         \includegraphics[width=\textwidth]{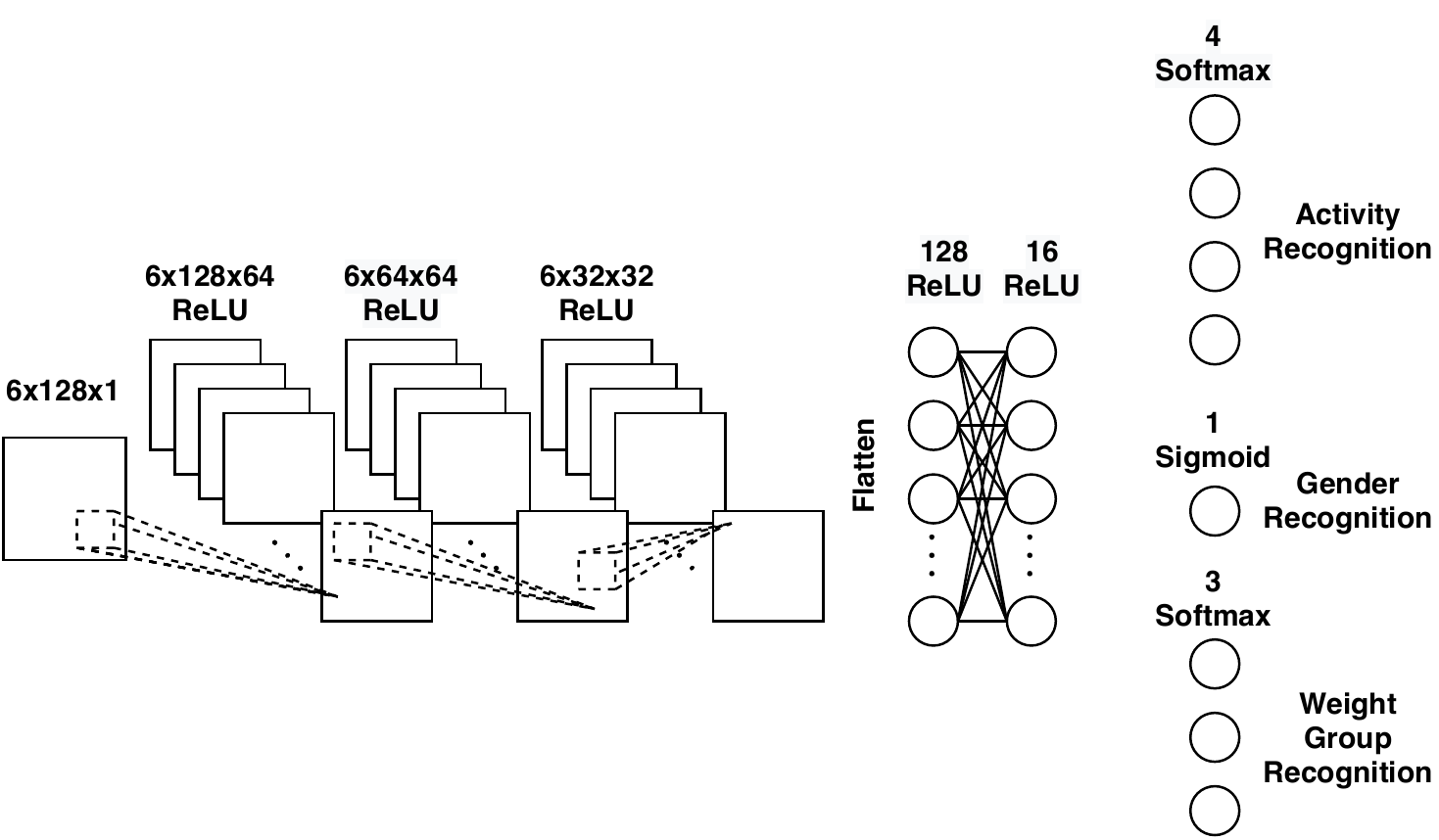}
         \caption{CNNs for detecting activity, gender, and weight group}
         \label{fig:mobi_cnn}
     \end{subfigure}
        \caption{Neural network architecture and the number of neurons in each layer for the MobiAct dataset.}
        \label{fig:three graphs4}
\end{figure}

Next consider the weight group anonymization task. \textcolor{red}{Similarly, on an average of three independent runs,}
we find that it takes $49.3$ and $49.8$ milliseconds to run the classifiers 
to identify the public and private attributes of one input data embedding, respectively.
The probabilistic encoder of the VAE model processes one embedding in $6.4$ milliseconds and the probabilistic decoder processes one embedding in $6.2$ milliseconds. Hence, it takes about $111.7$ milliseconds in total to anonymize one data embedding in the worst case. 
These results are shown in Tables~\ref{fig:mobiact_gen} and~\ref{fig:mobiact_weight} where the maximum running time is printed in bold in each case. Given that a new data embedding is generated every $500$ milliseconds, we conclude that data anonymization can be performed in real time on a typical edge device. The CNNs used for activity, gender, and weight classification, and the VAE architecture are depicted in Figures~\ref{fig:vae_mobi} and \ref{fig:mobi_cnn}.

\begin{table}[htbp]
\caption{The running time of different components of the proposed anonymization technique in the gender anonymization task in MobiAct.}
\begin{tabular}{c|c|c|c|c|c}
Model                & Batch Sizes & Type & nb. Embeddings & Time (s) & \begin{tabular}[c]{@{}c@{}}Time/\\ Embedding (s)\end{tabular}       \\ \toprule
Activity Recognition & 256         & CNN  & 20,301        & 1,002.4698 & \textbf{0.04938} \\
Gender Recognition   & 256         & CNN  & 20,301
    & 1,000.7468 & \textbf{0.04930} \\
Prob. Encoder 0      & 256         & MLP  & 9,563         & 58.8672  & 0.00616               \\
Prob. Decoder 0      & 256         & MLP  & 9,563
     & 57.0127   & 0.00596 \\
Prob. Encoder 1      & 256         & MLP  & 9,604         & 59.9439  & 0.00624                \\
Prob. Decoder 1      & 256         & MLP  & 9,604     & 59.0914  & 0.00615                \\
Prob. Encoder 2      & 256         & MLP  & 892           & 5.8286  & \textbf{0.00653}                \\
Prob. Decoder 2      & 256         & MLP  & 892           & 5.1610  & 0.00579                \\
Prob. Encoder 3      & 256         & MLP  & 255
         & 1.6024  & 0.00628 \\
Prob. Decoder 3      & 256         & MLP  & 255            & 1.6110   & \textbf{0.00632}                \\ \bottomrule
\end{tabular}
\label{fig:mobiact_gen}
\end{table}

\begin{table}[htbp]
\caption{The running time of different components of the proposed anonymization technique in the weight-group anonymization task in MobiAct.}
\begin{tabular}{c|c|c|c|c|c}
Model                & Batch Sizes & Type & nb. Embeddings & Time (s) & \begin{tabular}[c]{@{}c@{}}Time/\\ Embedding (s)\end{tabular}       \\ \toprule
Activity Recognition & 256         & CNN  & 20,301        & 1,001.1016 & \textbf{0.04931} \\
Weight Group Recognition   & 256         & CNN  & 20,301   & 1,011.6178 & \textbf{0.04983} \\
Prob. Encoder 0      & 256         & MLP  & 9,563         & 56.08   & 0.00586                \\
Prob. Decoder 0      & 256         & MLP  & 9,563         & 58.2625   & 0.00609 \\
Prob. Encoder 1      & 256         & MLP  & 9,641      & 53.9154   & 0.00559             \\
Prob. Decoder 1      & 256         & MLP  & 9,641         & 55.0189   & 0.00571                \\
Prob. Encoder 2      & 256         & MLP  & 914           & 5.8195    & \textbf{0.00637}                \\
Prob. Decoder 2      & 256         & MLP  & 914           & 5.6309   & \textbf{0.00616}                \\
Prob. Encoder 3      & 256         & MLP  & 252            & 1.4627   & 0.00580 \\
Prob. Decoder 3      & 256         & MLP  & 252            & 1.2952 & 0.00514                \\ \bottomrule
\end{tabular}
\label{fig:mobiact_weight}
\end{table}

\subsection{Comparison with Adversarial Model-based Anonymization Techniques}
For the sake of comparison and to get a better understanding of 
the overhead of obfuscating sensor data using our technique,
we measure the running time of the Replacement Autoencoder (RAE)~\cite{malekzadeh2017replacement} 
and the Anonymizing Autoencoder (AAE)~\cite{malekzadeh19}.
Recall that AAE is an adversarial model-based anonymization technique,
which was used as a baseline in Section~\ref{eval}.
We use the implementation of the methods that is available at~\cite{pmc_malek} and run it on our Raspberry Pi.
In this implementation, an embedding is passed through the Replacement Autoencoder before it is sent to the Anonymizing Autoencoder. 

We only discuss the results for the MotionSense dataset as these techniques are not originally used in~\cite{malekzadeh19} to anonymize data from the MobiAct dataset. Table~\ref{fig:malekz} shows the performance results. It can be seen that the total running time of their anonymization techniques is greater than 
the running time of our anonymization technique, which is $5.63$ milliseconds per embedding.
We conclude that our technique is not computationally expensive 
when it is compared to these adversarial model-based anonymization techniques.

\begin{table}[htbp]
\caption{The running time of two anonymization techniques proposed in the literature
in the gender anonymization task in MotionSense.}
\begin{tabular}{c|c|c|c|c|c}
Model                & Batch Sizes & Type & nb. Embeddings & Time (s) & \begin{tabular}[c]{@{}c@{}}Time/\\ Embedding (s)\end{tabular}       \\ \toprule
Replacement Autoencoder & 128         & CNN  & 13,873         & 62.23   & 0.0044857               \\
Anonymizing Autoencoder & 128         & CNN  & 13,873         & 199.79   & 0.0144013 \\
\bottomrule
\end{tabular}
\label{fig:malekz}
\end{table}

\subsection{Anonymization on Edge Devices versus Anonymization in the Cloud}
Partitioning neural network models and offloading parts of the computation 
to the cloud can decrease the running time and energy consumption at the edge~\cite{jeong2018ionn, neurosurgeon}.
Thus, we explore the possibility of sending the raw sensor data, latent space representations, 
or intermediate data to the cloud to reduce the running time of the anonymization application.
Clearly, sending raw sensor data to a remote server that performs 
anonymization can expose user data to the attacker.
Hence, the only part of the obfuscator that can run in the cloud is the probabilistic decoder. 
But since the running time of the probabilistic decoder \textcolor{red}{in the worst-case scenario} is around $2$ millisecond on a Raspberry Pi,
it does not make sense to run it in the cloud as the propagation delay is usually much greater than 
the running time of the decoder on the edge device.
Nevertheless, if an energy constraint is imposed on the edge device,
we may have to partition the probabilistic decoder and offload computation.
We plan to investigate this in future work.
\section{Conclusion}
\label{conc}

The number of IoT devices is estimated to surpass 20 billions worldwide by 2025.
Many of these devices are currently installed in our homes and workplaces, 
collecting significant amounts of data that can reveal private aspects of our lives 
if analyzed using advanced machine learning techniques.
Nevertheless, most users weigh privacy risks against the perceived benefits of IoT devices
and are reluctant to adopt privacy mechanisms that noticeably reduce these benefits.
Thus, it is crucial to develop data obfuscation techniques that enable the users 
to utilize the available data to its fullest potential without compromising their privacy.

In this paper, we extended the adversarial model-free anonymization technique
that we originally developed in~\cite{hajihass2020latent}. 
To achieve better anonymization performance, 
we augmented the loss function of the attribute-specific VAEs 
with the cross-entropy loss of a simple classification layer for the private attribute.
This encourages learning useful latent representations 
that better represent private attribute classes that exist in the dataset,
which turns out to be a useful property when we apply the transformation to the latent representations.
\textcolor{red}{Note that incorporating this classification layer, 
which is trained simultaneously with the VAE,
is different from using a model for adversarial training.
Instead of trying to fool a specific model, 
the classification layer is used merely to introduce useful structure into the latent space.}

The proposed adversarial model-free technique utilizes attribute-specific VAE models rather than a single VAE. 
This helps to reduce the overall size of the model and supports real-time anonymization at the edge. 
We evaluated our adversarial model-free anonymization technique 
with deterministic and probabilistic modifications on two HAR datasets, 
and corroborated that it outperforms the baseline, 
which is an adversarial model-based anonymization technique.
Furthermore, it is not as vulnerable to the re-identification attack as the baseline.


Due to the growing importance of real-time anonymization on edge devices, 
we studied the feasibility of adversarial model-free anonymization on a Raspberry~Pi~3 Model~B. 
We showed that the proposed adversarial model-free technique is capable of satisfying the time budget, 
and therefore, can be used in real time.
Moreover, we investigated the possibility of performing anonymization at the edge and in the cloud,
and found that partitioning neural network models does not make sense in terms of the running time, 
given the thread model considered in this work.
We plan to further explore this in future work.
\textcolor{red}{Another avenue for future work is to investigate how each user 
can navigate the privacy-utility tradeoff according to their needs and 
how the anonymization technique can be customized for every user.}

\bibliography{biblio} 
\bibliographystyle{ACM-Reference-Format}

\end{document}